\documentclass{article}

\PassOptionsToPackage{numbers, compress}{natbib}
\usepackage[final]{neurips_2021}

\usepackage[utf8]{inputenc} % allow utf-8 input
\usepackage[T1]{fontenc}    % use 8-bit T1 fonts
\usepackage{xcolor}         % colors
\usepackage[pagebackref=true,breaklinks=true,letterpaper=true,colorlinks,bookmarks=false]{hyperref}       % hyperlinks
\usepackage{url}            % simple URL typesetting
\usepackage{booktabs}       % professional-quality tables
\usepackage{amsfonts}       % blackboard math symbols
\usepackage{nicefrac}       % compact symbols for 1/2, etc.
\usepackage{microtype}      % microtypography
\usepackage[pdftex]{graphicx}
\usepackage{multirow}
\usepackage{enumitem}
\usepackage{wrapfig}
\usepackage{caption}
\usepackage{subcaption}
\usepackage{amssymb}
\usepackage{amsmath}

\makeatletter
\newcommand\figcaption{\def\@captype{figure}\caption}
\newcommand\tabcaption{\def\@captype{table}\caption}
\makeatother

\newcommand\blfootnote[1]{%
  \begingroup
  \renewcommand\thefootnote{}\footnote{#1}%
  \addtocounter{footnote}{-1}%
  \endgroup
}

\def\onedot{. }
\def\eg{\emph{e.g}\onedot}

\def\etc{\emph{etc}\onedot} 
 
\def\etal{\emph{et al}\onedot}

\title{History Aware Multimodal Transformer for Vision-and-Language Navigation}

\author{Shizhe Chen, Pierre-Louis Guhur, Cordelia Schmid, Ivan Laptev\\[3pt]
Inria, \'Ecole normale sup\'erieure, CNRS, PSL Research University\\ [3pt]
{\small \texttt{\{shizhe.chen, pierre-louis.guhur, cordelia.schmid, ivan.laptev\}@inria.fr}}\\ [3pt]
{\small \url{https://cshizhe.github.io/projects/vln_hamt.html}}}

\begin{document}

\maketitle

\begin{abstract}
Vision-and-language navigation (VLN) aims to build autonomous visual agents that follow instructions and navigate in real scenes. To remember previously visited locations and actions taken, most approaches to VLN implement memory using recurrent states. Instead, we introduce a History Aware Multimodal Transformer (HAMT) to incorporate a long-horizon history into multimodal decision making. HAMT efficiently encodes all the past panoramic observations via a hierarchical vision transformer (ViT), which first encodes individual images with ViT, then models 
spatial relation between images in a panoramic observation and finally takes into account temporal relation between panoramas in the history. It, then, jointly combines text, history and current observation to predict the next action. We first train HAMT end-to-end using several proxy tasks including single step action prediction and spatial relation prediction, and then use reinforcement learning to further improve the navigation policy. HAMT achieves new state of the art on a broad range of VLN tasks, including VLN with \emph{fine-grained instructions} (R2R, RxR), \emph{high-level instructions} (R2R-Last, REVERIE), \emph{dialogs} (CVDN) as well as \emph{long-horizon VLN} (R4R, R2R-Back). We demonstrate HAMT to be particularly effective for navigation tasks with longer trajectories. 
% We evaluate our model on three challenging VLN tasks, namely \emph{fine-grained VLN} (R2R), \emph{long-horizon VLN} (R4R and R2R-Back) and \emph{VLN with high-level instructions} (R2R-Last and REVERIE). HAMT outperforms the state of the art and achieves excellent results in all tasks. We also demonstrate HAMT to be particularly effective for navigation tasks  with longer trajectories. 
\end{abstract}

\section{Introduction}
\label{sec:intro}

% 1. introduce VLN task
Vision-and-language navigation (VLN) has recently received growing attention~\cite{anderson2018evaluation,chen2019touchdown,jain2019stay,zhang2020diagnosing,hong2020recurrent}. 
VLN requires an agent to understand natural language instructions, perceive the visual world, and  perform navigation actions to arrive at a target location.
A number of datasets have been proposed to support various VLN tasks such as indoor and outdoor navigation with fine-grained instructions \cite{chen2019touchdown,anderson2018vision,ku2020room}, language-driven remote object finding \cite{qi2020reverie} and navigation in dialogs \cite{thomason2020vision}.

% 2. challenge 1) history encoding
VLN agents are faced with several challenges.
First, as opposed to static vision-text grounding \cite{yu2018mattnet}, the agent continuously receives new visual observations and should align them with instructions. 
Most of existing works adopt recurrent neural networks (RNNs) \cite{anderson2018vision,fried2018speaker,tan2019learning,ma2019self,wang2019reinforced,hong2020language,lin2021scene} to encode historical observations and actions within a fixed-size state vector to predict the next action. Such condensed states might be sub-optimal for capturing essential information in extended trajectories~\cite{fang2019scene}.
For instance, \emph{``bring the spoon to me''} requires the agent to remember its start location after navigating to the \emph{``spoon''}, while early memories are prone to fade in the recurrent state. 
Few endeavors \cite{deng2020evolving,wang2021structured} construct external map-like memories for received observations. Nevertheless, these approaches still rely on RNNs to track the navigation state.
As the history plays an important role in environment understanding and instruction grounding, we propose to explicitly encode the history as a sequence of previous actions and observations instead of using recurrent states.

% 3. challenge 2) representations learning 
Another VLN challenge concerns the generalizations of agents to new environments that have not been observed during training~\cite{zhang2020diagnosing}.
One direction is to learn more generic text-image representations. 
The PRESS model \cite{li2019robust} improves language representation with a pretrained BERT encoder \cite{devlin2019bert}, and PREVALENT \cite{hao2020towards} uses pairs of instruction and single-step observations to pretrain a multimodal transformer. 
Though achieved promising results, these works do not optimize visual representation for the target navigation task. 
Moreover, lack of history in training~\cite{hao2020towards} makes it hard to learn cross-modal alignment and increases the risk of overfitting to training environments.
%more prone to memorize the structure of seen environments.
Another direction towards better generalization is to overcome exposure bias \cite{ranzato2015sequence} due to discrepancy between training and inference.
Different methods have been adopted for VLN including DAgger \cite{anderson2018vision,ross2011reduction} and scheduled sampling \cite{li2019robust,bengio2015scheduled}. Reinforcement Learning (RL) \cite{tan2019learning,sutton2018reinforcement} is one of the most effective approach among them, but it is considered unstable to directly train large-scale transformers via RL \cite{parisotto2020stabilizing}.

\begin{figure}
	\centering
	\includegraphics[width=.99\linewidth]{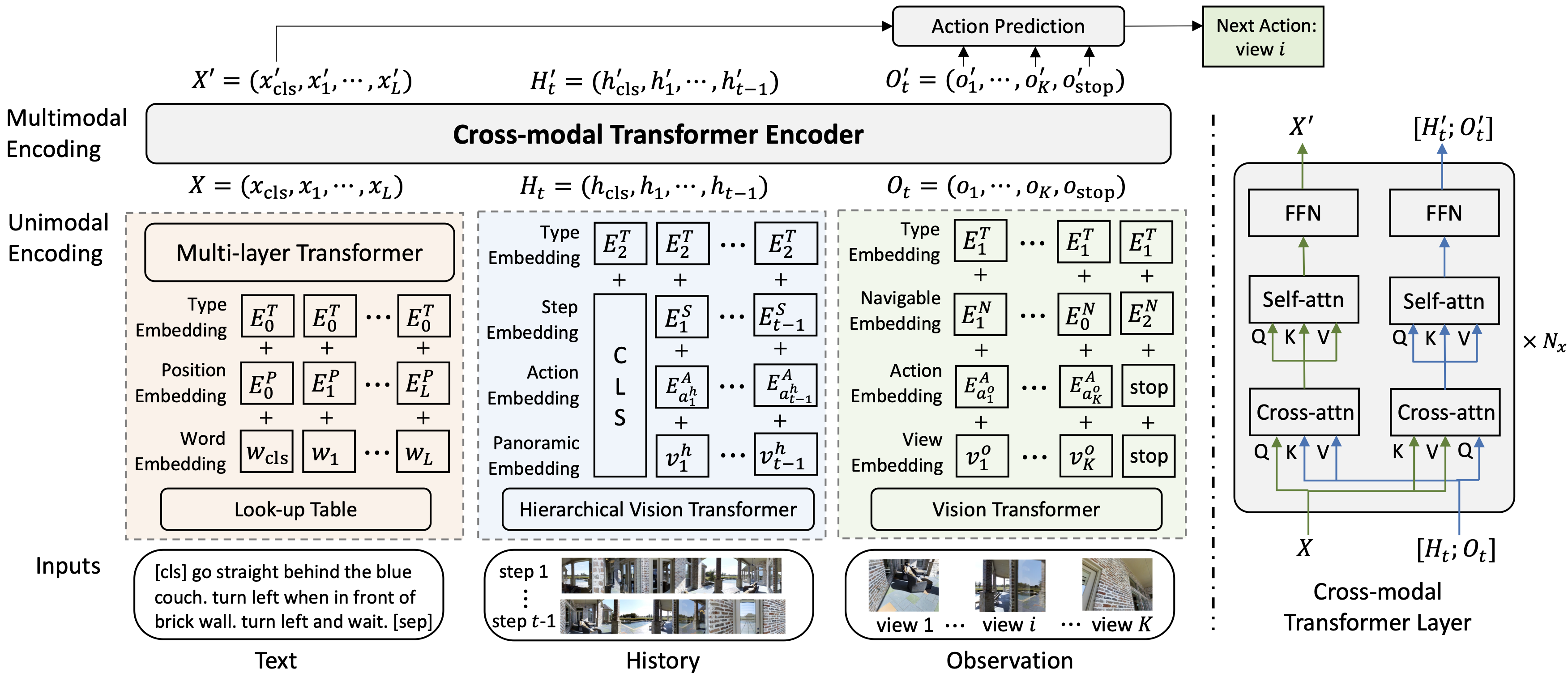}
	\caption{\small The architecture of History Aware Multimodal Tranformer (HAMT). HAMT jointly encodes textual instruction, full history of previous observations and actions, and current observation to predict the next action.}
	\label{fig:model_architecture}
% 	\vspace{-1.em}
\end{figure}

\begin{figure}
      \centering
      \begin{subfigure}[b]{0.7\textwidth}
      	\centering
      	\includegraphics[width=\linewidth]{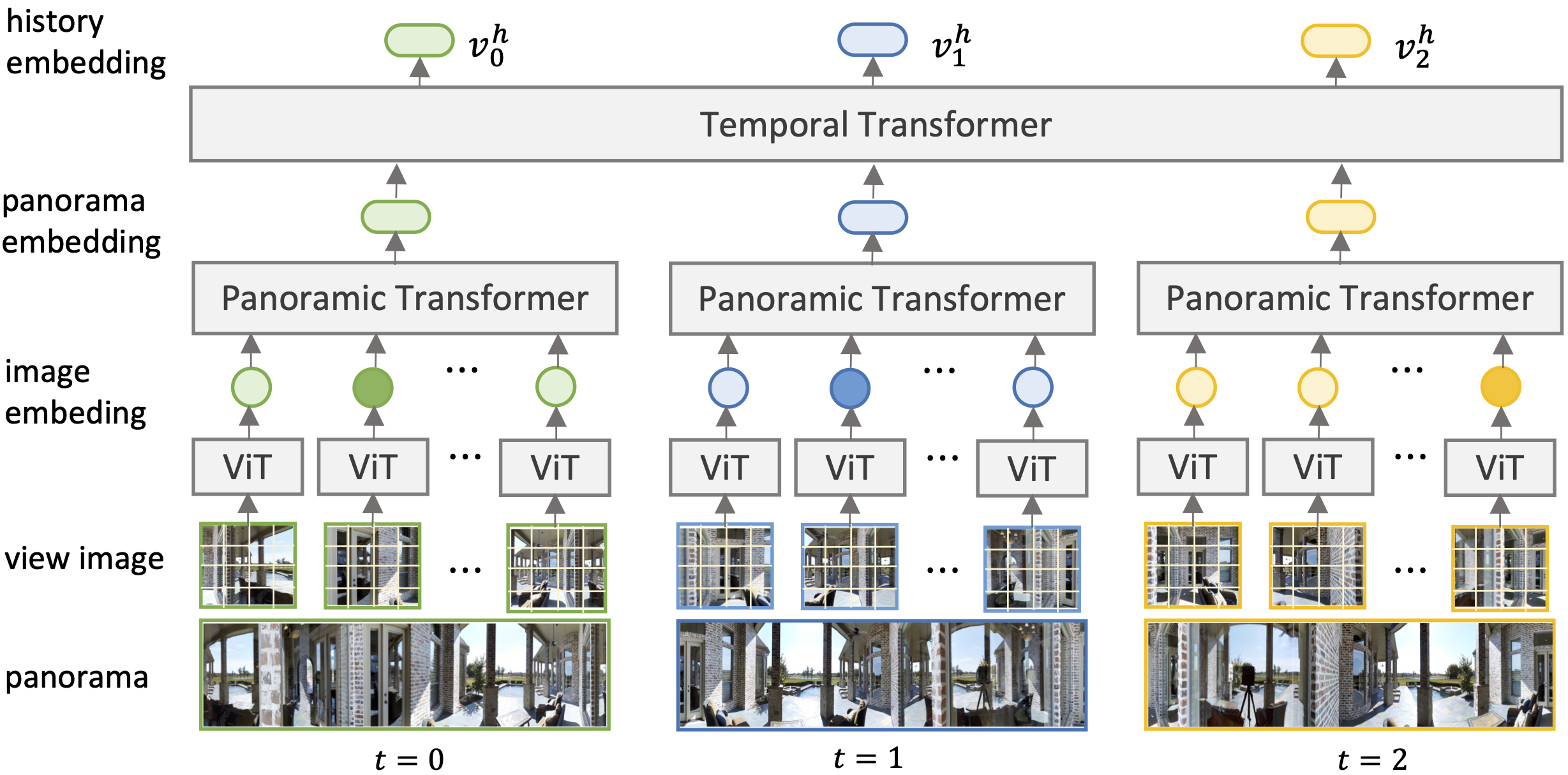}
      	\caption{\textbf{Hierarchical history encoding}. It first encodes individual view images with ViT, then models the spatial relation between images in each panorama, and finally captures the temporal relation between panoramas in the history.}
      	\label{fig:history_encoding_hierarchical}
      \end{subfigure}
      \hfill
      \begin{subfigure}[b]{0.28\textwidth}
        \centering
      	\includegraphics[width=0.7\linewidth]{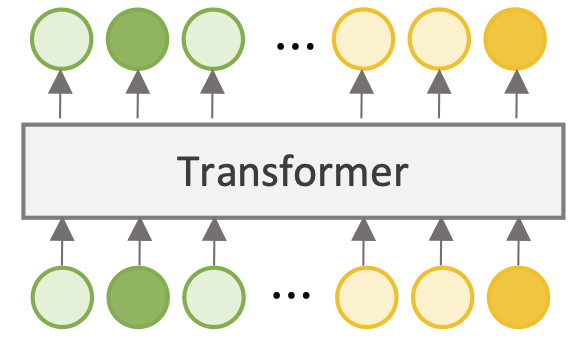}
      	\caption{\textbf{Flattened history encoding}. It encodes spatial and temporal relations at the same time.}
      	\label{fig:history_encoding_flatten}
      		
        \centering
        \includegraphics[width=0.7\linewidth]{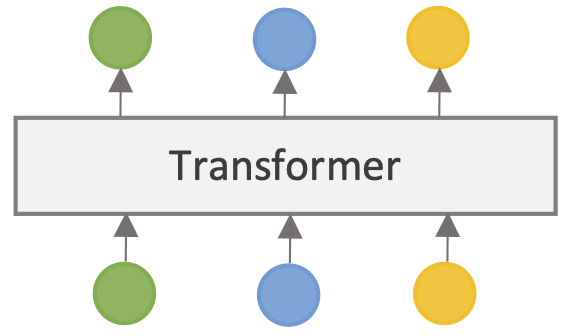}
        \caption{\textbf{Temporal-only history encoding}. It only considers temporal relation of oriented views.}
        \label{fig:history_encoding_tepmoral_only}
      \end{subfigure}
     \caption{ A comparison of history encoding methods. Circle nodes in different colors denote view images of panorama at different steps. Darker circle nodes are the oriented view of the agent.}
     \label{fig:history_encoding}
    %  \vspace{-1em}
 \end{figure}

% 4. our proposed approach
To address the above challenges, we propose the History Aware Multimodal Transformer (HAMT), a fully transformer-based architecture for multimodal decision making in VLN tasks.
As illustrated in Figure~\ref{fig:model_architecture}, HAMT consists of unimodal transformers for text, history and observation encoding, and a cross-modal transformer to capture long-range dependencies of the history sequence, current observation and instruction.
Since our history contains a sequence of all previous observations, its encoding is computationally expensive.
To resolve complexity issues, we propose a hierarchical vision transformer as shown in Figure~\ref{fig:history_encoding}, which progressively learns representations for a single view, spatial relationships among views within a panorama and, finally, the temporal dynamics across panoramas of the history.
In order to learn better visual representations, we propose auxiliary proxy tasks for end-to-end training. Such tasks include single-step action prediction based on imitation learning, self-supervised spatial relationship reasoning, masked language and image predictions and instruction-trajectory matching.
We empirically show that our training facilitates the subsequent fine-tuning of our model with RL~\cite{mnih2016asynchronous}.
We carry out extensive experiments on various VLN tasks, including VLN with \emph{fine-grained instructions} (R2R \cite{anderson2018vision} and RxR \cite{ku2020room}), \emph{high-level instructions} (REVERIE \cite{qi2020reverie} and our proposed R2R-Last), \emph{dialogs} \cite{thomason2020vision} as well as \emph{long-horizon VLN} (R4R \cite{jain2019stay} and our proposed R2R-Back which requires the agent to return back after arriving at the target location).
HAMT outperforms state of the art on both seen and unseen environments in all the tasks. 

We summarize our contributions as follows: 
(1)~We introduce HAMT to efficiently model long-horizon history of observed panoramas and actions via hierarchical vision transformer;
(2)~We train HAMT with auxiliary proxy tasks in an end-to-end fashion and use RL to improve the navigation policy;
(3)~We validate our method and outperform state of the art in a diverse range of VLN tasks, while demonstrating larger gains for long-horizon navigation.

\section{Related work}
\label{sec:related_work}

\noindent\textbf{Vision-and-language navigation.}
Training instruction-following navigation agents has attracted increasing research attention \cite{anderson2018evaluation,chen2019touchdown,anderson2018vision,ku2020room,qi2020reverie,shridhar2020alfred}.
% lstm models: cross-modal alignment
Anderson \etal \cite{anderson2018vision} propose a sequence-to-sequence LSTM baseline for the VLN task.
Fried \etal \cite{fried2018speaker} extend it with panoramic action space and synthesized instructions.
To improve cross-modal alignment, the self-monitoring agent \cite{ma2019self} proposes co-grounding and progress estimation, and RelGraph \cite{hong2020language} uses graphs to model relationships across scene, objects and directions.
% rl methods
Reinforcement learning (RL) is typically used to improve navigation policy.
The EnvDrop model \cite{tan2019learning} mixes imitation learning and A3C \cite{mnih2016asynchronous}.
The RCM \cite{wang2019reinforced} utilizes intrinsic reward of cross-modal matching in REINFORCE algorithm.
Wang \etal \cite{wang2020soft} propose to learn rewards via soft expert distillation.
% transformer models
Due to the success of transformer \cite{vaswani2017attention}, recent works explore transformer architectures in VLN.
PRESS \cite{li2019robust} replaces LSTM instruction encoder with pretrained BERT \cite{devlin2019bert}.
SIA \cite{lin2021scene} uses transformer for single-step multimodal fusion and LSTM for sequential action prediction.
PTA \cite{landi2019perceive} is a transformer VLN model using CNNs to extract visual features~\cite{he2016deep}. 
Here we propose the first full transformer architecture for VLN and train it end-to-end.

\noindent\textbf{Memory-based policy for navigation.}
% lstm recurrent memory
LSTMs \cite{hochreiter1997long} have been the dominant approach to encode memories for navigation \cite{anderson2018vision,fried2018speaker,tan2019learning,wang2019reinforced}.
Condensing all history into one feature vector, however, is prone to the loss of information. 
% topological memory
Alternative approaches include topological map memory structures~\cite{gupta2017cognitive,savinov2018semi}.
Deng \etal \cite{deng2020evolving} use graphs to capture environment layout and enable long-term planing.
A similar graph is adopted in \cite{wang2021structured} with frontier-exploration based decision making.
But these works still utilize LSTMs for state tracking.
% transformer memory
To exploit long-term spatio-temporal dependencies, Fang \etal \cite{fang2019scene} store histories in a sequence encoded with transformer.
Recurrent VLN-BERT \cite{hong2020recurrent} injects a recurrent unit to encode histories in transformer for VLN.
The most similar work to ours is Episodic Transformer (E.T.)~\cite{pashevich2021episodic}. Differently from~\cite{pashevich2021episodic}, we propose a hierarchical encoding of the panoramic observation history and optimize the whole model in end-to-end training. 

\noindent\textbf{Multimodal pretraining with transformers.}
Recent works show significant progress in vision and language tasks using multimodal pretraining.
In particular, transformer architectures such as one-stream \cite{chen2020uniter,li2020oscar} and dual-stream \cite{lu2019vilbert,tan2019lxmert} achieve state of the art for a number of downstream tasks including visual question answering, image-text retrieval and image captioning.
While most previous methods rely on CNN to extract image representations, ViLT~\cite{kim2021vilt} adopts Vision Transformer (ViT)~\cite{dosovitskiy2020image} and trains it with associated texts in an end-to-end manner thanks to the efficiency of ViT.
A few endeavors \cite{hao2020towards,majumdar2020improving} explore multimodal pretraining for VLN.
PREVALENT~\cite{hao2020towards} pretrains a transformer using instructions and single-step observations without referring to trajectory history.
VLN-BERT~\cite{majumdar2020improving} measures the compatibility between an instruction and images in a path but does not support action prediction.
Our work presents the first end-to-end trainable VLN transformer that jointly encodes text, history and observation, and is able to sequentially predict actions.

\section{Method}
\label{sec:method}

\paragraph{Problem definition}
The VLN problem \cite{anderson2018vision} is formulated as a partially observable Markov decision process, where future observations are independent of the past conditioning on current state $s_t$.
Given an instruction $\mathcal{W}$ containing a sequence of $L$ words $(w_1, w_2, \cdots, w_{L})$, an agent should follow the instruction to move in a connectivity graph to reach the goal location.
At each step $t$, the agent receives an observation $\mathcal{O}_t$, a panorama of its surrounding environment. The $\mathcal{O}_t$ consists of $K$ single view images split from the panorama $\mathcal{O}_t \triangleq ([v^o_1; a^o_1], \cdots, [v^o_{K};a^o_K])$, where $v^o_i$ is the visual feature of the $i$-th view and $a^o_i$ denotes the relative angle to face the view (subscript $t$ is omitted for simplicity).
There are $n$ navigable viewpoints among all the $K$ views\footnote{A navigable view can lead to one or multiple viewpoints. We follow \cite{hong2020recurrent,tan2019learning} to use different features for these viewpoints. The viewpoints share the same visual features but differ in angle features.}, denoted as $\mathcal{O}_t^c \triangleq ([v_1^c;a_1^c], \cdots, [v_n^c;a_n^c])$.
We follow the setup in \cite{fried2018speaker} and use $\mathcal{O}_t^c$ as the decision space, so the agent only needs to select a candidate in $\mathcal{O}_t^c$ at each step.
All observations $\mathcal{O}_i$ and performed actions $a_i^h$ before step $t$ form the history $\mathcal{H}_t \triangleq ([\mathcal{O}_1; a^h_1], \cdots, [\mathcal{O}_{t-1}; a^{h}_{t-1}])$, where $a_i^h$ denotes the turned angles at step $i$.
The goal is to learn a policy $\pi$ parametrized by $\Theta$ to predict the next action based on the instruction, history and the current observation, which is $\pi(a_t | \mathcal{W}, \mathcal{H}_t, \mathcal{O}_t, \mathcal{O}_t^c; \Theta)$.

Unlike dominant recurrent approaches to condense $\mathcal{H}_t$ into a fixed-size vector, in this section, we present the History Aware Multimodal Transformer (HAMT) that jointly encodes text,  long-horizon history, and observation for sequential action prediction.
The model architecture is described in Section~\ref{sec:transformer_model}. %As large-scale transformers are hard to optimize in RL objectives \cite{parisotto2020stabilizing}, 
We propose end-to-end training for HAMT in Section~\ref{sec:model_pretrain} to learn unimodal and multimodal representations, and then use RL to fine-tune the navigation policy in Section~\ref{sec:model_finetune}.

\subsection{HAMT: History Aware Multimodal Transformer}
\label{sec:transformer_model}
Figure~\ref{fig:model_architecture} illustrates the model architecture of HAMT. 
The inputs text $\mathcal{W}$, history $\mathcal{H}_t$ and observation $\mathcal{O}_t$ are first encoded via the corresponding unimodal transformers respectively, and then fed into the cross-modal transformer encoder to capture multimodal relationships.

\noindent\textbf{Text Encoding.}
For each token $i$ in the instruction $\mathcal{W}$, we embed it as the summation of its word embedding $w_i$, position embedding $E^P_i$ and type embedding of text $E^T_0$. 
Then we employ a transformer with $N_L$ layers to obtain contextual representation $x_i$ following the standard BERT \cite{devlin2019bert}.

\noindent\textbf{Observation Encoding.}
For each view $[v^o_i;a^o_i]$ in the panoramic observation $\mathcal{O}_t$, we first represent the relative angle $a^o_i$ as $E^A_{a^o_i} = (\sin \theta_i, \cos \theta_i, \sin \phi_i, \cos \phi_i)$ where $\theta_i$ and $\phi_i$ are the relative heading and elevation angle to the agent's orientation. Then the observation embedding $o_i$ is as follows:
\begin{equation}
	\label{eqn:obs_embed}
	o_i = \mathrm{LN}(W^o_v v^o_i) + \mathrm{LN}(W^o_a E^A_{a^o_i}) + E^N_{o_i} +  E^T_{1}
\end{equation}
where $W^o_v, W^o_a$ are learnable weights. The $E^N_{o_i}$ denotes the navigable embedding to differentiate types of views, with $E^N_0$ for non-navigable view,  $E^N_1$ for navigable view and $E^N_2$ for stop view (we append a stop token in observation to support stop action). The $E^T_{1}$ is the type embedding of observation. We omit bias terms for simplicity.
The $\mathrm{LN}$ denotes layer normalization \cite{ba2016layer}. Because $a^o_i$ has much lower feature dimensions than $v^o_i$, we apply $\mathrm{LN}$ to balance the encoded $a^o_i$ and $v^o_i$.

\noindent\textbf{Hierarchical History Encoding.}
As $\mathcal{H}_t$ consists of all the past panoramic observations $\mathcal{O}_{i}$ and performed actions $a^h_{i}$ before step $t$, it is important to encode $\mathcal{H}_t$ efficiently as context.
Figures~\ref{fig:history_encoding_flatten}-\ref{fig:history_encoding_tepmoral_only} depict the flattened and temporal-only history encoding approaches used in VLN-BERT \cite{majumdar2020improving} and E.T. \cite{pashevich2021episodic} respectively.
The flattened approach treats each view image in $\mathcal{O}_{i}$ as a token, so the history sequence contains $tK$ tokens. Though it enables to learn relationships among all image views, the computation cost quadratically increases with the sequence length, making it inefficient for long-horizon tasks. 
In the temporal-only approach, only the oriented view of the agent in each $\mathcal{O}_{i}$ is taken as inputs instead of the whole panorama, so only $t$ temporal tokens are encoded. However, this approach can lose critical information in past observations. For example, in the instruction \emph{``with the windows on your left, walk through the large room past the sitting areas''}, the object  \emph{``window''} does not appear in the oriented view of the agent. Therefore, the encoded history is insufficient to tell whether the agent passed the window or not, making the model confused to take the next action.
%it is hard to align the trajectory with the instruction correctly and accumulate sufficient knowledge of the environment.

In order to balance computational efficiency and information integrity, we propose a hierarchical history encoding approach as illustrated in Figure~\ref{fig:history_encoding_hierarchical}. It hierarchically encodes view images within each panorama and then temporal relationships across panoramas, similar to the factorized spatial-temporal video transformer \cite{arnab2021vivit}.
For each $\mathcal{O}_i$, its constituent view images are first embeded via ViT and Eq~(\ref{eqn:obs_embed}), and then encoded via a panoramic transformer with $N_h$ layers to learn spatial relationships within the panorama.
We apply average pooling to obtain panorama embedding, and add it with the oriented view image feature in residual connection. 
The parameters in ViT and panoramic transformer are shared for different steps.
In this way, each historical observation $\mathcal{O}_i$ is represented as $v^h_i$, and the final temporal token $h_i$ is computed as:
\begin{equation}
	h_i = \mathrm{LN}(W^h_v v^h_i) + \mathrm{LN}(W^h_a E^A_{a^h_i}) + E^S_i + E^T_2
\end{equation}
where $E^S_i$ denotes the $i$-th step embedding, $E^T_2$ is the type embedding of history.
The computational cost is $O(t K^2 + t^2)$, which significantly reduces from $O(t^2 K^2)$ in the flattened approach.
To be noted, we add a special token \verb|[cls]| to the start of the history sequence to obtain a global representation.
The embedding of \verb|[cls]| is a parameter to learn, which is initialized from a zero vector. 

\noindent\textbf{Cross-modal Encoding.}
We concatenate history and observation as the vision modality, and use cross-modal transformer with $N_x$ layers to fuse features from text, history and observation as shown in the right of Figure~\ref{fig:model_architecture}.
The reason of using such dual-stream architecture rather than one-stream is that the length of different modalities can be highly imbalanced, and the dual-stream architecture can balance the importance of intra- and inter-modal relationships by model design~\cite{cao2020behind}.
In each cross-modal layer, a vision-text cross-attention is firstly performed for vision modality to attend relevant text information and vice versa for text modality. Then each modality uses self-attention to learn intra-modal relationship such as interaction between observation and history, followed by a fully-connected neural network.
Finally, the HAMT model outputs embeddings $X^{'}=(x'_{\mathrm{cls}}, x'_{1}, \cdots, x'_{L}), H_t^{'}=(h'_{\mathrm{cls}}, h'_1, \cdots, h'_{t-1}), O^{'}_t=(o'_1, \cdots, o'_{K}, o'_{\mathrm{stop}})$ for tokens in text, history and observation respectively.

\subsection{End-to-end training with proxy tasks}
\label{sec:model_pretrain}
As it is difficult to train large-scale transformers with RL due to sparse supervision \cite{parisotto2020stabilizing}, we propose to first end-to-end train HAMT via several proxy tasks to learn unimodal and multimodal representation.

Table~\ref{tab:pretrain_cmpr} compares our HAMT with previous VLN transformers PREVALENT~\cite{hao2020towards} and VLN-BERT~\cite{majumdar2020improving} in inputs and proxy tasks.
As neither PREVALENT nor VLN-BERT jointly encodes text, history and observation, a limited choice of proxy tasks can be applied in training.
Our model instead can take advantage of various proxy tasks to learn cross-modal alignment, spatial and temporal reasoning, and history-aware action prediction.
Given the input pair $(\mathcal{W}, \mathcal{H}_T)$ where $T$ is the length of full trajectory, we can apply common proxy tasks as in vision-and-language pretraining \cite{lu2019vilbert,majumdar2020improving}, including Masked Language Modeling (MLM), Masked Region Modeling (MRM) and Instruction Trajectory Matching (ITM). Details of the three proxy tasks are presented in the supplementary material.
In the following, we introduce new proxy tasks given the triplet input $(\mathcal{W}, \mathcal{H}_t, \mathcal{O}_t)$ specifically for VLN tasks. 

\begin{table}
	\small
	\centering
	\caption{Comparison of HAMT and previous VLN transformers.}
	\label{tab:pretrain_cmpr}
	\begin{tabular}{ccccccccc} \toprule
		\multirow{2}{*}{Models} & \multicolumn{3}{c}{Inputs} & \multicolumn{5}{c}{Proxy Tasks} \\
		& Text & History & Observation & MLM & MRM & ITM & SAP/SAR & SPREL \\ \midrule
		PREVALENT \cite{hao2020towards} & \checkmark &  & \checkmark & \checkmark &  &  & \checkmark &  \\
		VLN-BERT \cite{majumdar2020improving} & \checkmark & \checkmark &  & \checkmark & \checkmark & \checkmark &  &  \\
		HAMT (Ours) & \checkmark & \checkmark & \checkmark & \checkmark & \checkmark & \checkmark & \checkmark &  \checkmark \\ \bottomrule
	\end{tabular}
% 	\vspace{-2em}
\end{table}

\noindent\textbf{Single-step Action Prediction/Regression (SAP/SAR).}
The task deploys imitation learning to predict the next action based on instruction, history from expert demonstration and the current observation.
We formulate it as a classification and a regression task respectively. 
In the SAP classification task, we predict action probability for each navigable view in $\mathcal{O}^c_t$ which is $p_t(o'_i) = \frac{\mathrm{exp}(f_{\text{SAP}}(o'_i \odot x'_{\mathrm{cls}}))}{\sum_{j} \mathrm{exp}(f_{\text{SAP}}(o'_j \odot x'_{\mathrm{cls}}))}$, where $f_{\text{SAP}}$ is a two-layer fully-connected network, $\odot$ is element-wise multiplication and $x'_\mathrm{cls}$ is output embedding of special text token \verb|[cls]|.
The objective is to minimize negative log probability of the target view action $o'_*$: $L_{\mathrm{SAP}} = - \mathrm{log}\ p_t(o'_*)$.
In SAR regression task, we directly predict the action heading and elevation angles based on the text token \verb|[cls]| which is $\hat{\theta_t}, \hat{\phi_t} = f_{\text{SAR}}(x'_{\mathrm{cls}})$.
The loss function is $L_{\mathrm{SAR}} = (\hat{\theta_t} - \theta_t)^2 + (\hat{\phi_t} - \phi_t)^2$.
The two proxy tasks enable the model to learn how to make action decision conditioning on instruction and contextual history.

\noindent\textbf{Spatial Relationship Prediction (SPREL).}
Expressions of egocentric and allocentric spatial relations are frequent in navigational instructions, such as \emph{``walk into the room on your left''} and \emph{``enter the bedroom next to the stairs''}.
In order to learn spatial relation aware representations, we propose the SPREL self-supervised task to predict relative spatial position of two views in a panorama based on only visual feature, angle feature or both.
Assume $[v^o_i;a^o_i]$ and $[v^o_j;a^o_j]$ are two views in $\mathcal{O}_t$, we randomly zero out $v^o_*$ or $a^o_*$ with probability of 0.3. Their encoded representations are $o'_i$ and $o'_j$, and their relative heading and elevation angles are $\theta_{ij}, \phi_{ij}$.
We then predict $\hat{\theta}_{ij}, \hat{\phi}_{ij} = f_{\text{SPREL}}([o'_i; o'_j])$ where $[;]$ denotes vector concatenation and optimize $L_{\mathrm{SPREL}} = (\hat{\theta}_{ij} - \theta_{ij})^2 + (\hat{\phi}_{ij} - \phi_{ij})^2$.
The task helps for spatial relationship reasoning in the observation.

\noindent\textbf{Training Strategy.}
Instead of directly training the whole HAMT model at once, we propose to progressively train HAMT in two stages. 
In the first stage, we freeze ViT pretrained on ImageNet \cite{deng2009imagenet} and train the rest of the modules which are randomly initialized. This aims to avoid catastrophic forgetting of the pretrained weights in ViT.
Then we unfreeze ViT and train the whole model end-to-end. The learning rate for ViT is set to be higher than for others modules to avoid vanishing gradients and to speedup convergence.
We empirically show that the proposed two-stage training outperforms one-stage training in the supplementary material.

\subsection{Fine-tuning for sequential action prediction}
\label{sec:model_finetune}
\noindent\textbf{Structure Variants.}
We present two variants of HAMT for action prediction in the following.
1)~MLP action head: we directly reuse the action prediction network $f_{\text{SAP}}$ in the SAP task to predict navigable views. We use it as default for VLN tasks.
2) MLP action head based on encoder-decoder structure: the original HAMT model applies cross-modal attention for both vision-to-text and text-to-vision, which is computationally expensive when instructions are long. Therefore, we remove the cross-modal attention from text to vision. In this way, we separate the cross-modal transformer into an encoder which only takes instruction as input, and a decoder that inputs history and observation as query and attends over encoded text tokens.
Please see supplementary material for details.

\noindent\textbf{RL+IL Objective.}
We combine Reinforcement Learning (RL) and Imitation Learning (IL) to fine-tune HAMT for sequential action prediction. The IL relies on the SAP loss defined in Section~\ref{sec:model_pretrain} and follows the expert action at each step while RL samples actions according to the policy $\pi$.
Specifically, we use the Asynchronous Advantage Actor-Critic (A3C) RL algorithm \cite{mnih2016asynchronous}.
At each step $t$, the agent samples an action based on policy $\pi$: $\hat{a}^h_t \sim \pi(a_t | \mathcal{W}, \mathcal{H}_t, \mathcal{O}_t, \mathcal{O}^c_t)$ and receives an immediate reward $r_t$.
For non-stop actions, we set $r_t$ as the reduced distance of taking the action to the target and the increased alignment score~\cite{jain2019stay} compared to expert demonstration as defined in~\cite{hong2020recurrent}; for the stop action, $r_t=2$ if the agent successfully arrives at the target otherwise -2.
A critic network is trained to estimate the value of each state $s_t$, which is $R_t=\sum_{k=0}^{T-t} \gamma^{k} r_{t+k}$ where $\gamma$ is discount factor. We implement it as $V_t=f_{\text{critic}}(x'_{\mathrm{cls}} \odot h'_{\mathrm{cls}})$. 
%The A3C updates the weights $\Theta$ of policy $\pi$ as $\Theta \leftarrow \Theta + \mu \frac{1}{T} \sum_{t=1}^{T}\nabla_{\Theta} \mathrm{log}\ \pi(\hat{a}^h_t; \Theta) (R_t - V_t)$, where $\mu$ is the learning rate. 
As the reward signal favors shortest distance, we empirically find it benefits to combine A3C RL with IL weighted by $\lambda$, which is:
\begin{equation}
    % \vspace{-1em}
    \label{eqn:finetune_objective}
	\Theta \leftarrow \Theta + \underbrace{\mu \frac{1}{T} \sum_{t=1}^{T}\nabla_{\Theta} \mathrm{log}\ \pi(\hat{a}^h_t; \Theta) (R_t - V_t)}_{\text{Reinforcement Learning (RL)}} + \underbrace{\lambda \mu \frac{1}{T^*} \sum_{t=1}^{T^*} \nabla_{\Theta} \text{log}\ \pi(a^*_t;\Theta)}_{\text{Imitation Learning (IL)}}
\end{equation}
where $\mu$ is the learning rate, $a^*_t$ is the expert action at step $t$ of the expert trajectory of length $T^*$.

\section{Experiments}
\label{sec:expr}

\subsection{Experimental setup}

\noindent\textbf{Datasets.}
We evaluate our method on four VLN tasks (seven datasets): \emph{VLN with fine-grained instructions} (R2R \cite{anderson2018vision}, RxR \cite{ku2020room}); \emph{VLN with high-level instructions} (REVERIE \cite{qi2020reverie}, R2R-Last); \emph{vision-and-dialogue navigation} (CVDN \cite{thomason2020vision}); and \emph{long-horizon VLN} (R4R \cite{jain2019stay}, R2R-Back).
\parskip=0.1em
\begin{itemize}[itemsep=0.1em,parsep=0em,topsep=0em,partopsep=0em,leftmargin=2em]
	\item {\bfseries R2R}~\cite{anderson2018evaluation} builds upon Matterport3D \cite{chang2017matterport3d} and includes 90 photo-realistic houses with 10,567 panoramas. It contains 7,189 shortest-path trajectories, each associated with 3 instructions. The dataset is split into train, val seen, val unseen and test unseen sets with 61, 56, 11 and 18 houses respectively. Houses in val seen split are the same as training, while houses in val unseen and test splits are different from training.
	\item {\bfseries RxR}~\cite{ku2020room} is a large multilingual VLN dataset based on Matterport 3D. The instructions are in three different languages (English, Hindi and Telugu). The dataset emphasizes the role of language in VLN by addressing biases in paths and describing more visible entities than R2R.
	\item {\bfseries R4R}~\cite{jain2019stay} extends R2R dataset by concatenating two adjacent tail-to-head trajectories in R2R. Therefore, it has longer instructions and trajectories. The trajectories are also less biased as they are not necessarily the shortest-path from start to end location.
	\item {\bfseries R2R-Back} is a new VLN setup proposed in this work. The agent is required to return to its start location after arriving at the destination. The agent needs to remember its navigation histories to solve the task. We add a return command at the end of each instruction in R2R and a reverse path from the end to start locations as expert demonstration. 
	\item {\bfseries CVDN}~\cite{thomason2020vision} defines a navigation from dialog history task, which requires an agent to arrive at goal regions based on multi-turn question-answering dialogs. Such types of instructions are often ambiguous and under-specified. The lengths of instructions and paths are also long.
	\item {\bfseries REVERIE}~\cite{qi2020reverie} replaces step-by-step instructions in R2R with high-level instructions, which mainly describe the target location and object. The agent, hence, is required to navigate to the goal without detailed guidance and depends on its past experiences.
	\item {\bfseries R2R-Last} is our proposed VLN setup similar to REVERIE. It only uses the last sentence from the original R2R instructions describing the final destination.
\end{itemize}

\noindent\textbf{Evaluation metrics.}
We adopt standard metrics \cite{anderson2018evaluation}, including (1) Trajectory Length (TL): the agent’s navigated path in meters; (2) Navigation Error (NE): the average distance in meters between the agent’s final position and the target; (3) Success Rate (SR): the ratio of trajectories reaching the destination with a maximum error of 3 meters to the target; and (4) Success Rate normalized by the ratio between the length of the shortest path and the predicted path (SPL). SPL is more relevant than SR as it balances the navigation accuracy and efficiency. 
% The SR on the R2R-Back dataset is defined as arriving at the original target and then successfully returning back to the start, otherwise an agent that never moves would achieve 100\% SR.
% The above metrics measure whether the agent successfully arrives at the target in shortest distance, however, trajectories in R4R and R2R-Back datasets are no more the shortest path between two locations.
For long-horizon VLN task (R4R and R2R-Back), we further employ three metrics to measure the path fidelity between the predicted path and target path, including (5) Coverage weighted by Length Score (CLS) \cite{jain2019stay}; (6) the normalized Dynamic Time Warping (nDTW) \cite{ilharco2019general}; and (7) the Success weighted by nDTW (SDTW).

\noindent\textbf{Implementation details.}
% details of architecture
For the HAMT model, we set $N_L=9$ for language transformer, $N_h=2$ for panoramic transformer in hierarchical history encoding, and $N_x=4$ for cross-modal transformer.
There are $K=36$ view images in each panoramic observation.
We use ViT-B/16 \cite{dosovitskiy2020image} for image encoding if not otherwise specified.
In training with proxy tasks, we randomly select proxy tasks for each mini-batch with predefined ratio. We train HAMT for 200k iterations with fixed ViT using learning rate of 5e-5 and batch size of 64 on 4 NVIDIA Tesla P100 GPUs ($\sim$1 day).
The whole HAMT model is trained end-to-end for 20k iterations on 20 NVIDIA V100 GPUs with learning rate of 5e-5 for ViT and 1e-5 for the others ($\sim$20 hours).
We use R2R training set and augmented pairs from \cite{hao2020towards} for training unless otherwise noted.
In fine-tuning with RL+IL, we set $\lambda=0.2$ in Eq~(\ref{eqn:finetune_objective}) and $\gamma=0.9$. 
The model is fine-tuned for 100k iterations with learning rate of 1e-5 and batch size of 8 on a single GPU. Unimodal encoders are fixed by default.
The best model is selected according to performance on val unseen split.
We use the same augmented data as \cite{hong2020recurrent} for R2R for fair comparison, while no augmented data is used for other datasets.
Greedy search is applied in inference following the single-run setting.
Please see supplementary material for more details. 

\subsection{Ablation studies}
In this section, we evaluate each component in the HAMT model, including: hierarchical history encoding, end-to-end training with proxy tasks, and fine-tuning objectives.

\paragraph{How important is the history encoding for VLN?}
\begin{wraptable}{r}{0.6\linewidth}
\centering
\small
\vspace{-1em}
\caption{R2R navigation results for alternative methods of history encoding. All methods use Resnet152 visual features and are trained from scratch on R2R dataset.}
\label{tab:hist_cmpr_from_scratch}
\begin{tabular}{ccccc} \toprule
\multirow{2}{*}{\begin{tabular}[c]{@{}c@{}}History \\ Encoding \end{tabular}} & \multicolumn{2}{c}{Val Seen} & \multicolumn{2}{c}{Val Unseen}  \\ 
 & SR$\uparrow$ & SPL$\uparrow$ & SR$\uparrow$ & SPL$\uparrow$ \\ \midrule
\begin{tabular}[c]{@{}c@{}}RecBERT \cite{hong2020recurrent} \\\end{tabular} & 62 & 59 & 50 & 46 \\ \midrule
Recurrent & 60.9$_{\pm 1.0}$ & 56.6$_{\pm 1.1}$ & 52.2$_{\pm 0.7}$ & 47.0$_{\pm 0.5}$  \\
Temporal-only & 61.5$_{\pm 0.8}$ & 57.7$_{\pm 0.7}$ & 53.2$_{\pm 0.1}$ & 48.0$_{\pm 0.4}$   \\
Hierarchical & \textbf{65.5}$_{\pm 1.2}$ & \textbf{61.3}$_{\pm 1.4}$ & \textbf{54.4}$_{\pm 0.4}$ & \textbf{48.7}$_{\pm 0.4}$  \\ \bottomrule
\end{tabular}
\vspace{-1em}
\end{wraptable}
% \footnotetext[1]{The result is reported in \cite{hong2020recurrent} using Resnet152 visual features and  OSCAR structure without initialization.}

For fair comparison with the state-of-the-art recurrent architecture RecBERT \cite{hong2020recurrent}, we use the same Resnet152 visual features and train all the models from scratch with RL+IL objectives to avoid the influence of different weight initialization.
The models are optimized for 300k iterations end-to-end except for the visual feature.
Table~\ref{tab:hist_cmpr_from_scratch} compares different history encoding approaches on R2R dataset.
Our recurrent model slightly differs from RecBERT (no init. OSCAR) \cite{hong2020recurrent} in transformer architecture as shown in Figure~\ref{fig:model_architecture}. It  achieves slightly better performance on val unseen split.
The temporal-only model uses transformer to encode agent's oriented visual observations in history sequence, and outperforms the recurrent method by relative gains of 1.9\% on SR and 2.1\% on SPL for val unseen split.
Adding panoramic observations in a hierarchical way results in 4.2\% (SR) and 3.6\% (SLP) relative improvements on the val unseen split compared to the recurrent method.
Even larger improvements are achieved on val seen split as the hierarchical model has a larger capacity to fit the seen environments.
This evaluation demonstrates the advantage of our hierarchical history representation compared to the recurrent and temporal-only history representation.

\begin{table}
    \caption{Ablations for end-to-end HAMT training on R2R dataset using proposed proxy tasks.}
     \label{tab:r2r_pretrain}
    \begin{subtable}[h]{0.5\textwidth}
        \centering
        \small
        \tabcolsep=0.06cm
        \caption{Comparison of visual features and end-to-end training. The ``PT'' stands for proxy tasks in training; ``e2e'' for optimizing the visual representation.}
        \label{tab:r2r_vit_e2e}
        \begin{tabular}{cccccccc} \toprule
        \multirow{2}{*}{feature} & \multirow{2}{*}{PT} & \multirow{2}{*}{e2e} & \multicolumn{2}{c}{Val Seen} & \multicolumn{2}{c}{Val Unseen} \\
        & & & SR$\uparrow$ & SPL$\uparrow$ & SR$\uparrow$ & SPL$\uparrow$ \\ \midrule
        \multirow{2}{*}{\begin{tabular}[c]{@{}c@{}}Resnet \\ 152\end{tabular}} & $\times$ & $\times$ & 65.5$_{\pm 1.2}$ & 61.3$_{\pm 1.4}$ & 54.4$_{\pm 0.4}$ & 48.7$_{\pm 0.4}$ \\
         & \checkmark & $\times$ & 69.3$_{\pm 1.0}$ & 64.8$_{\pm 1.2}$ & 63.5$_{\pm 0.5}$ & 57.5$_{\pm 0.5}$ \\ \midrule
        \multirow{2}{*}{ViT} & \checkmark & $\times$ & \textbf{75.7}$_{\pm 1.0}$ & \textbf{72.5}$_{\pm 1.0}$ & 64.4$_{\pm 0.3}$ & 58.8$_{\pm 0.0}$ \\
         & \checkmark & \checkmark & 75.0$_{\pm 0.9}$ & 71.7$_{\pm 0.7}$ & \textbf{65.7}$_{\pm 0.7}$ & \textbf{60.9}$_{\pm 0.7}$ \\ 
        \bottomrule
        \end{tabular}
    \end{subtable}
    \hfill
    \begin{subtable}[h]{0.45\textwidth}
        \small
        \tabcolsep=0.06cm
        \caption{Comparison of different proxy tasks. The ``SAP(R)'' denotes the single step action prediction and regression task, and ``SPREL'' is the spatial relationship prediction task.}
        \label{tab:r2r_proxy_tasks}
        \begin{tabular}{ccccccc} \toprule
        \multirow{2}{*}{\begin{tabular}[c]{@{}c@{}}SAP \\ (R) \end{tabular}} & \multirow{2}{*}{\begin{tabular}[c]{@{}c@{}}SP \\ REL \end{tabular}} & \multicolumn{2}{c}{Val Seen} & \multicolumn{2}{c}{Val Unseen} \\
        & & SR$\uparrow$ & SPL$\uparrow$ & SR$\uparrow$ & SPL$\uparrow$ \\ \midrule
        $\times$ & $\times$ & 71.2$_{\pm 2.3}$ & 67.2$_{\pm 2.0}$ & 62.8$_{\pm 1.3}$ & 57.7$_{\pm 1.0}$ \\ \midrule
        \checkmark & $\times$ & 74.7$_{\pm 0.6}$ & 71.1$_{\pm 0.9}$ & 63.6$_{\pm 0.1}$ & 58.1$_{\pm 0.4}$ \\
        \checkmark & \checkmark & \textbf{75.7}$_{\pm 1.0}$ & \textbf{72.5}$_{\pm 1.0}$ & \textbf{64.4}$_{\pm 0.3}$ & \textbf{58.8}$_{\pm 0.0}$\\ \bottomrule
        \end{tabular}
     \end{subtable}
     \vspace{-1em}
\end{table}

\paragraph{How much does training with proxy tasks help?}
We next evaluate the advantage of training HAMT end-to-end with proxy tasks.
In Table~\ref{tab:r2r_vit_e2e}, the first row uses RL+IL objectives to train HAMT from scratch, while the second row uses proxy tasks for training prior to RL+IL fine-tuning.
We can see that it significantly boosts the performance to first train with proxy tasks. It improves on val unseen split with 16.7\% and 18.0\% relative gains on SR and SPL respectively, indicating that training with auxiliary proxy tasks enables better generalization.
In the third row, we replace the visual feature from Resnet152 to ViT. The ViT feature improves the performance on both val seen and val unseen splits, showing that more powerful visual representations matter.
Finally, training ViT end-to-end obtains 2.1\% gains on SPL on val unseen split. 
This is the first time to show that optimizing visual representations end-to-end is beneficial for VLN tasks.
In Table~\ref{tab:r2r_proxy_tasks}, we evaluate the benefit of the two new proxy tasks for frozen ViT features using the other proxy tasks by default.
The SAP(R) uses imitation learning to predict actions, which directly influences the navigation policy and improves the performance by a large margin.
The SPREL is a self-supervised proxy task that forces the model to learn spatial relationships in panorama and helps generalization in unseen environments.
More experiments to ablate contributions from history encoding and proxy tasks, contributions of proxy tasks in end-to-end training \etc are presented in supplementary material.

\begin{wraptable}{r}{0.48\textwidth}
	\centering
	\small
	\tabcolsep=0.1cm
	\vspace{-1em}
	\caption{Ablations for fine-tuning objectives of sequential action prediction on R2R dataset.}
	\label{tab:r2r_finetune_rl}
	\begin{tabular}{ccccccc} \toprule
		\multirow{2}{*}{IL} & \multirow{2}{*}{RL}& \multicolumn{2}{c}{Val Seen} & \multicolumn{2}{c}{Val Unseen} \\
		& & SR$\uparrow$ & SPL$\uparrow$ & SR$\uparrow$ & SPL$\uparrow$ \\ \midrule
		$\times$ & $\times$ & 57.9 & 54.8 & 51.8 & 48.9 \\
		\checkmark & $\times$ & 63.7$_{\pm 2.1}$ & 61.7$_{\pm 2.2}$ & 57.2$_{\pm 0.1}$ & 54.7$_{\pm 0.3}$ \\
		$\times$ & \checkmark & 70.5$_{\pm 2.9}$ & 65.6$_{\pm 2.8}$ & 63.5$_{\pm 1.4}$ & 57.5$_{\pm 1.1}$ \\
		\checkmark & \checkmark & \textbf{75.0}$_{\pm 0.9}$ & \textbf{71.7}$_{\pm 0.7}$ & \textbf{65.7}$_{\pm 0.7}$ & \textbf{60.9}$_{\pm 0.7}$ \\
		\bottomrule
	\end{tabular}
	\vspace{-1em}
\end{wraptable}

\paragraph{What is the impact of the fine-tuning objectives?}
Table~\ref{tab:r2r_finetune_rl} presents results using different objectives in fine-tuning. 
The first row directly applies HAMT trained by proxy tasks, which achieves lower performance than that after IL fine-tuning, because we mainly use augmented data in proxy task training to increase visual diversity, but such noisy data deteriorates action prediction performance.
Previous work \cite{tan2019learning} has shown that RL alone performs poorly. However, training with proxy tasks stabilizes the followup RL fine-tuning. HAMT optimized by RL achieves much better performance than that when fine-tuning with IL on the SR metric. 
It indicates that RL is able to learn better exploration strategy on unseen environments.
However, as the reward for RL focuses more on shortest paths rather than path fidelity with instructions, the improvement on SPL metric is relatively small compared to SR metric. Moreover, the fluctuation of the pure RL objective is larger than IL.
Therefore, mixing the RL and IL achieves the best performance.

\subsection{Comparison to state of the art}

\paragraph{VLN with fine-grained instructions: R2R and RxR.}
\begin{table}
	\centering
	\small
	\tabcolsep=0.15cm
% 	\vspace{-1.5em}
	\caption{Comparison with state-of-the-art methods on R2R dataset.}
	\label{tab:r2r_sota_cmpr}
	\begin{tabular}{lcccccccccccc} \toprule
		\multirow{2}{*}{Methods} & \multicolumn{4}{c}{Validation Seen} & \multicolumn{4}{c}{Validation Unseen} & \multicolumn{4}{c}{Test Unseen} \\
		& TL & NE$\downarrow$ & SR$\uparrow$ & SPL$\uparrow$ & TL & NE$\downarrow$ & SR$\uparrow$ & SPL$\uparrow$ & TL & NE$\downarrow$ & SR$\uparrow$ & SPL$\uparrow$ \\ \midrule
		Seq2Seq \cite{anderson2018vision} & 11.33 & 6.01 & 39 & - & 8.39 & 7.81 & 22 & - & 8.13 & 7.85 & 20 & 18 \\
		SF \cite{fried2018speaker} & - & 3.36 & 66 & - & - & 6.62 & 35 & - & 14.82 & 6.62 & 35 & 28 \\
		PRESS \cite{li2019robust}& 10.57 & 4.39 & 58 & 55 & 10.36 & 5.28 & 49 & 45 & 10.77 & 5.49 & 49 & 45 \\
		EnvDrop \cite{tan2019learning} & 11.00 & 3.99 & 62 & 59 & 10.70 & 5.22 & 52 & 48 & 11.66 & 5.23 & 51 & 47 \\
		AuxRN \cite{zhu2020vision} & - & 3.33 & 70 & 67 & - & 5.28 & 55 & 50 & - & 5.15 & 55 & 51 \\
		PREVALENT \cite{hao2020towards} & 10.32 & 3.67 & 69 & 65 & 10.19 & 4.71 & 58 & 53 & 10.51 & 5.30 & 54 & 51 \\
		RelGraph \cite{hong2020language}& 10.13 & 3.47 & 67 & 65 & 9.99 & 4.73 & 57 & 53 & 10.29 & 4.75 & 55 & 52 \\
		RecBERT \cite{hong2020recurrent} & 11.13 & 2.90 & 72 & 68 & 12.01 & 3.93 & 63 & 57 & 12.35 & 4.09 & 63 & 57 \\ \midrule
		HAMT (Ours) & 11.15 & \textbf{2.51} & \textbf{76} & \textbf{72} & 11.46 & \textbf{3.62} & \textbf{66} & \textbf{61} & 12.27 & \textbf{3.93} & \textbf{65} & \textbf{60} \\ \bottomrule
		%HAMT (Ours) & 10.94 & 2.70 & 75 & 72 & 11.87 & 3.65 & 65 & 59 & 12.65 & 4.11 & 63 & 58 \\ \bottomrule
	\end{tabular}
	\vspace{-1em}
\end{table}

% \begin{table}
% \centering
% \small
% \tabcolsep=0.15cm
% \caption{Comparison with state-of-the-art methods on R2R dataset.}
% \label{tab:r2r_sota_cmpr}
% \begin{tabular}{lcccccccccccc} \toprule
% \multirow{2}{*}{Methods} & \multicolumn{4}{c}{Validation Seen} & \multicolumn{4}{c}{Validation Unseen} & \multicolumn{4}{c}{Test Unseen} \\
% & TL & NE$\downarrow$ & SR$\uparrow$ & SPL$\uparrow$ & TL & NE$\downarrow$ & SR$\uparrow$ & SPL$\uparrow$ & TL & NE$\downarrow$ & SR$\uparrow$ & SPL$\uparrow$ \\ \midrule
% Seq2Seq \cite{anderson2018vision} & 11.33 & 6.01 & 39 & - & 8.39 & 7.81 & 22 & - & 8.13 & 7.85 & 20 & 18 \\
% SF \cite{fried2018speaker} & - & 3.36 & 66 & - & - & 6.62 & 35 & - & 14.82 & 6.62 & 35 & 28 \\
% PRESS \cite{li2019robust}& 10.57 & 4.39 & 58 & 55 & 10.36 & 5.28 & 49 & 45 & 10.77 & 5.49 & 49 & 45 \\
% EnvDrop \cite{tan2019learning} & 11.00 & 3.99 & 62 & 59 & 10.70 & 5.22 & 52 & 48 & 11.66 & 5.23 & 51 & 47 \\
% AuxRN \cite{zhu2020vision} & - & 3.33 & 70 & 67 & - & 5.28 & 55 & 50 & - & 5.15 & 55 & 51 \\
% PREVALENT \cite{hao2020towards} & 10.32 & 3.67 & 69 & 65 & 10.19 & 4.71 & 58 & 53 & 10.51 & 5.30 & 54 & 51 \\
% RelGraph \cite{hong2020language}& 10.13 & 3.47 & 67 & 65 & 9.99 & 4.73 & 57 & 53 & 10.29 & 4.75 & 55 & 52 \\
% RecBERT \cite{hong2020recurrent} & 11.13 & 2.90 & 72 & 68 & 12.01 & 3.93 & 63 & 57 & 12.35 & 4.09 & 63 & 57 \\ \midrule
% THOR (Ours) & 10.94 & 2.70 & 75 & 72 & 11.87 & 3.65 & 65 & 59 & 12.65 & 4.11 & 63 & 58 \\ \bottomrule
% \end{tabular}
% \end{table}

Table~\ref{tab:r2r_sota_cmpr} compares HAMT with previous VLN methods on the R2R benchmark.
Our model outperforms state-of-the-art results of RecBERT~\cite{hong2020recurrent} by relative 5.9\% and 7.0\% improvements in SPL on val seen and unseen splits respectively.
We achieve state-of-the-art performance under the single-run setting on the unseen testing split of the leaderboard\footnote{We report the published results on the testing unseen split as shown in  \url{https://eval.ai/web/challenges/challenge-page/97/leaderboard/270} (25/10/2021).}.
It demonstrates the effectiveness and generalization of our model.
We further provide computation time in inference for HAMT and RecBERT in the supplementary material to show the efficiency of our HAMT model.
We also achieve large improvements on RxR dataset. The full results are presented in supplementary material.

\paragraph{Long-horizon VLN: R4R and R2R-Back.}
Table~\ref{tab:r4r_sota_cmpr} shows navigation results on R4R dataset.
As R4R contains longer instructions and trajectories compared to R2R, we use the encoder-decoder variant of HAMT for better efficiency.
Our method outperforms previous approaches in all metrics and shows particularly large improvements for the path fidelity related metrics.
Compared to RecBERT, HAMT achives 8.2\% and 9.5\% relative improvement in CLS and nDTW respectively. The large improvements on these path fidelity related metrics indicate that HAMT is better to follow the designated path of the fine-grained instruction.
Figure~\ref{fig:r4r_instr_lens} evaluates the performance of HAMT and RecBERT with respect to instruction length measured by words. Though the nDTW decreases for longer instructions, the relative improvement of HAMT increases with the instruction length.

\begin{minipage}{0.55\textwidth}
    \small
    \vspace{0.5em}
    \tabcolsep=0.13cm
    \tabcaption{Comparison on R4R val unseen split.}
    \label{tab:r4r_sota_cmpr}
    \begin{tabular}{lccccccccccc} \toprule
    	Methods & NE$\downarrow$ & SR$\uparrow$ & CLS$\uparrow$ & nDTW$\uparrow$ & SDTW$\uparrow$ \\ \midrule
    	SF \cite{fried2018speaker} & 8.47 & 24 & 30 & - & -  \\
		RCM \cite{wang2019reinforced} & - & 29 & 35 & 30 & 13 \\
    	PTA \cite{landi2019perceive} & 8.25 & 24 & 37 & 32 & 10  \\
    	EGP \cite{deng2020evolving} & 8.0 & 30.2& 44.4 & 37.4 & 17.5 \\
    % 	SSM \cite{wang2021structured} & 8.27 & 32 & 53 & 39 & 19 \\
		RelGraph \cite{hong2020language} & 7.43 & 36 & 41 & 47 & 34 \\
    	RecBERT$^{\dagger}$ \cite{hong2020recurrent} & 6.67 & 43.6 & 51.4 & 45.1 & 29.9 \\ \midrule
    	HAMT (Ours) & \textbf{6.09} & \textbf{44.6} & \textbf{57.7} & \textbf{50.3} & \textbf{31.8}  \\ \bottomrule
	\end{tabular}
	\vspace{0.8em}
\end{minipage}
\begin{minipage}{.4\textwidth}
\centering
\vspace{0.8em}
\includegraphics[width=0.85\linewidth]{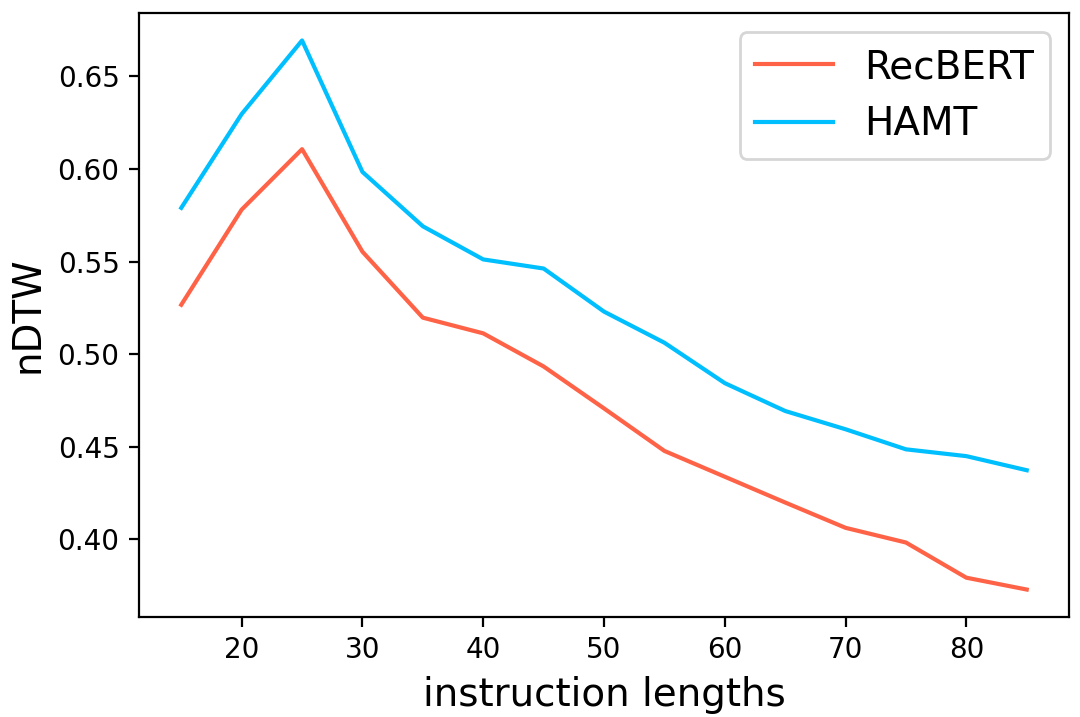}
\figcaption{nDTW with respect to instruction length on R4R val unseen split.}
\label{fig:r4r_instr_lens}
\vspace{0.5em}
\end{minipage}

The navigation performance on R2R-Back dataset is presented in Table~\ref{tab:r2rback_sota_cmpr}. 
We compare with two state-of-the-art recurrent models EnvDrop \cite{tan2019learning} and RecBERT \cite{hong2020recurrent} based on LSTM and transformer respectively (both models are trained on R2R-Back for fair comparison).
The improvements are more significant on this task as it requires the agent to remember the way it came to the target to successfully return back. The recurrent state is insufficient to capture such history and leads to inferior performance compared to the HAMT model.

\begin{table}[h]
	\centering
	\small
	\tabcolsep=0.18cm
	\vspace{-1.2em}
	\caption{Comparison of methods on the R2R-Back dataset.}
	\label{tab:r2rback_sota_cmpr}
	\begin{tabular}{lcccccccccc} \toprule
		\multirow{2}{*}{Methods} & \multicolumn{5}{c}{Val Seen} & \multicolumn{5}{c}{Val Unseen} \\ 
	    & TL & SR$\uparrow$ & SPL$\uparrow$  & nDTW$\uparrow$  & SDTW$\uparrow$ & TL & SR$\uparrow$  & SPL$\uparrow$  & nDTW$\uparrow$  & SDTW$\uparrow$  \\ \midrule
		EnvDrop$^{\dagger}$ \cite{tan2019learning} & 23.83 & 44.1 & 42.0 & 61.3 & 39.4 & 24.57 & 32.4 & 30.2 & 51.1 & 28.0 \\
		RecBERT$^{\dagger}$ \cite{hong2020recurrent} & 22.33 & 51.4 & 48.4 & 67.3 & 45.7 & 23.35 & 41.1 & 37.7 & 58.2 & 35.6 \\ \midrule
		HAMT (Ours) & 22.76 & \textbf{64.8} & \textbf{61.8} & \textbf{73.7} & \textbf{58.9} & 23.78 & \textbf{57.2} & \textbf{53.1} & \textbf{65.1} & \textbf{49.5} \\ \bottomrule
	\end{tabular}
	\vspace{-1em}
\end{table}

\paragraph{Vision-and-Dialog Navigation: CVDN.}

\begin{wraptable}{l}{0.55\textwidth}
\centering
\small
\tabcolsep=0.08cm
\caption{Navigation performance on CVDN dataset.}
\label{tab:cvdn_sota_cmpr}
\begin{tabular}{lccc} \toprule
 & Val Seen & Val Unseen & Test Unseen \\ \midrule
PREVALENT \cite{hao2020towards} & - & 3.15 & 2.44 \\
VISITRON \cite{shrivastava2021visitron} & 5.11 & 3.25 & 3.11 \\
MT-RCM+EnvAg \cite{wang2020environment} & 5.07 & 4.65 & 3.91 \\ \midrule
HAMT (Ours) & \textbf{6.91} & \textbf{5.13} & \textbf{5.58} \\ \bottomrule
\end{tabular}
\end{wraptable}

The CVDN dataset contains dialogs as instructions and use Goal Progress (GP) in meters as the primary evaluation metric. GP measures the difference between completed distance and left distance to the goal, so the higher the better.
There are two types of demonstrations in the dataset. One is shortest-path trajectory and the other is player's navigation trajectory.
We mix the two types of demonstrations as supervision in training which has shown to be the most effective in previous works \cite{hao2020towards,shrivastava2021visitron,wang2020environment}.
As navigation paths in CVDN dataset are much longer than R2R dataset, we adopt the encoder-decoder variant of HAMT.
As shown in Table~\ref{tab:cvdn_sota_cmpr}, HAMT outperforms existing recurrent approaches on both seen and unseen environments, and achieves the top position in the leaderboard\footnote{\url{https://eval.ai/web/challenges/challenge-page/463/leaderboard/1292} (25/10/2021)}.
It demonstrates that our HAMT model is generalizable to different types of instructions in new VLN tasks.

\paragraph{VLN with high-level instructions: R2R-Last and REVERIE.}

\begin{wraptable}{r}{.5\textwidth}
\centering
\small
\vspace{-1em}
	\caption{Comparison on the R2R-Last dataset.}
	\label{tab:r2rlast_sota_cmpr}
	\begin{tabular}{lcccc} \toprule
		\multirow{2}{*}{Methods} & \multicolumn{2}{c}{Val Seen} & \multicolumn{2}{c}{Val Unseen} \\
		& SR$\uparrow$ & SPL$\uparrow$ & SR$\uparrow$ & SPL$\uparrow$  \\ \midrule
		EnvDrop$^{\dagger}$ \cite{tan2019learning} & 42.8 & 38.4 &  34.3 & 28.3    \\
		RecBERT$^{\dagger}$ \cite{hong2020recurrent} & 50.2 & 45.8 & 41.6 & 37.3  \\ \midrule
		HAMT (Ours) & \textbf{53.3} & \textbf{50.3} & \textbf{45.2} & \textbf{41.2}  \\ \bottomrule
	\end{tabular}
\end{wraptable}
Table~\ref{tab:r2rlast_sota_cmpr} shows results on the R2R-Last dataset that specifies the goal location and contains no step-by-step instructions.
The HAMT model with the hierarchical history encoding is able to better accumulate the knowledge of the environment and achieves 9.8\% and 10.5\% relative gains on SPL metric on seen and unseen splits respectively compared to RecBERT \cite{hong2020recurrent}\blfootnote{$^\dagger$ We run the original implementation of methods released by the authors.}.
The REVERIE dataset also contains high-level instructions but requires object grounding at the target location besides navigation. We provide results on REVERIE dataset in supplementary material. Our HAMT achieves SPL 30.20 and 26.67 on val unseen and test splits respectively, outperforming the state of the art navigation performance \cite{hong2020recurrent} by 5.3\% and 2.7\%.

\section{Conclusion}
This paper presents the first end-to-end transformer for vision-and-language navigation, denoted as History Aware Multimodal Transformer (HAMT). Our method efficiently encodes long-horizon history and combines it with instructions and observations to derive multimodal action prediction. 
The HAMT is first trained with proxy tasks in an end-to-end manner, and is then fine-tuned with RL to improve the navigation policy.
We achieve state-of-the-art navigation performance on a diverse range of challenging VLN tasks, demonstrating improved accuracy and generalization of our approach compared to the dominant recurrent methods.
% broader impact
Future work could extend our history-aware transformer to VLN with continuous actions \cite{krantz2020beyond} and could benefit from pretraining on larger navigation datasets. 
This paper has minimal ethical, privacy and safety concerns.

\begin{ack}
This work was granted access to the HPC resources of IDRIS under the allocation 101002 made by GENCI. 
It was funded in part by the French government under management of Agence Nationale de la Recherche as part of the ``Investissements d'avenir'' program, reference ANR19-P3IA-0001 (PRAIRIE 3IA Institute) and by Louis Vuitton ENS Chair on Artificial Intelligence.
\end{ack}

%\bibliographystyle{unsrt}
%\bibliography{reference}

\begin{thebibliography}{10}

\bibitem{anderson2018evaluation}
Peter Anderson, Angel Chang, Devendra~Singh Chaplot, Alexey Dosovitskiy,
  Saurabh Gupta, Vladlen Koltun, Jana Kosecka, Jitendra Malik, Roozbeh
  Mottaghi, Manolis Savva, et~al.
\newblock On evaluation of embodied navigation agents.
\newblock {\em arXiv preprint arXiv:1807.06757}, 2018.

\bibitem{chen2019touchdown}
Howard Chen, Alane Suhr, Dipendra Misra, Noah Snavely, and Yoav Artzi.
\newblock Touchdown: Natural language navigation and spatial reasoning in
  visual street environments.
\newblock In {\em Proceedings of the IEEE/CVF Conference on Computer Vision and
  Pattern Recognition}, pages 12538--12547, 2019.

\bibitem{jain2019stay}
Vihan Jain, Gabriel Magalhaes, Alexander Ku, Ashish Vaswani, Eugene Ie, and
  Jason Baldridge.
\newblock Stay on the path: Instruction fidelity in vision-and-language
  navigation.
\newblock In {\em Proceedings of the 57th Annual Meeting of the Association for
  Computational Linguistics}, pages 1862--1872, 2019.

\bibitem{zhang2020diagnosing}
Yubo Zhang, Hao Tan, and Mohit Bansal.
\newblock Diagnosing the environment bias in vision-and-language navigation.
\newblock {\em International Joint Conferences on Artificial Intelligence},
  2020.

\bibitem{hong2020recurrent}
Yicong Hong, Qi~Wu, Yuankai Qi, Cristian Rodriguez-Opazo, and Stephen Gould.
\newblock Vln bert: A recurrent vision-and-language bert for navigation.
\newblock In {\em Proceedings of the IEEE/CVF Conference on Computer Vision and
  Pattern Recognition}, pages 1643--1653, 2021.

\bibitem{anderson2018vision}
Peter Anderson, Qi~Wu, Damien Teney, Jake Bruce, Mark Johnson, Niko
  S{\"u}nderhauf, Ian Reid, Stephen Gould, and Anton Van Den~Hengel.
\newblock Vision-and-language navigation: Interpreting visually-grounded
  navigation instructions in real environments.
\newblock In {\em Proceedings of the IEEE Conference on Computer Vision and
  Pattern Recognition}, pages 3674--3683, 2018.

\bibitem{ku2020room}
Alexander Ku, Peter Anderson, Roma Patel, Eugene Ie, and Jason Baldridge.
\newblock Room-across-room: Multilingual vision-and-language navigation with
  dense spatiotemporal grounding.
\newblock In {\em Proceedings of the 2020 Conference on Empirical Methods in
  Natural Language Processing (EMNLP)}, pages 4392--4412, 2020.

\bibitem{qi2020reverie}
Yuankai Qi, Qi~Wu, Peter Anderson, Xin Wang, William~Yang Wang, Chunhua Shen,
  and Anton van~den Hengel.
\newblock Reverie: Remote embodied visual referring expression in real indoor
  environments.
\newblock In {\em Proceedings of the IEEE/CVF Conference on Computer Vision and
  Pattern Recognition}, pages 9982--9991, 2020.

\bibitem{thomason2020vision}
Jesse Thomason, Michael Murray, Maya Cakmak, and Luke Zettlemoyer.
\newblock Vision-and-dialog navigation.
\newblock In {\em Conference on Robot Learning}, pages 394--406. PMLR, 2020.

\bibitem{yu2018mattnet}
Licheng Yu, Zhe Lin, Xiaohui Shen, Jimei Yang, Xin Lu, Mohit Bansal, and
  Tamara~L Berg.
\newblock Mattnet: Modular attention network for referring expression
  comprehension.
\newblock In {\em Proceedings of the IEEE Conference on Computer Vision and
  Pattern Recognition}, pages 1307--1315, 2018.

\bibitem{fried2018speaker}
Daniel Fried, Ronghang Hu, Volkan Cirik, Anna Rohrbach, Jacob Andreas,
  Louis-Philippe Morency, Taylor Berg-Kirkpatrick, Kate Saenko, Dan Klein, and
  Trevor Darrell.
\newblock Speaker-follower models for vision-and-language navigation.
\newblock In {\em Proceedings of the 32nd International Conference on Neural
  Information Processing Systems}, pages 3318--3329, 2018.

\bibitem{tan2019learning}
Hao Tan, Licheng Yu, and Mohit Bansal.
\newblock Learning to navigate unseen environments: Back translation with
  environmental dropout.
\newblock In {\em Proceedings of the 2019 Conference of the North American
  Chapter of the Association for Computational Linguistics: Human Language
  Technologies, Volume 1 (Long and Short Papers)}, pages 2610--2621, 2019.

\bibitem{ma2019self}
Chih-Yao Ma, Jiasen Lu, Zuxuan Wu, Ghassan AlRegib, Zsolt Kira, Richard Socher,
  and Caiming Xiong.
\newblock Self-monitoring navigation agent via auxiliary progress estimation.
\newblock In {\em Proceedings of the International Conference on Learning
  Representations}, 2019.

\bibitem{wang2019reinforced}
Xin Wang, Qiuyuan Huang, Asli Celikyilmaz, Jianfeng Gao, Dinghan Shen,
  Yuan-Fang Wang, William~Yang Wang, and Lei Zhang.
\newblock Reinforced cross-modal matching and self-supervised imitation
  learning for vision-language navigation.
\newblock In {\em Proceedings of the IEEE/CVF Conference on Computer Vision and
  Pattern Recognition}, pages 6629--6638, 2019.

\bibitem{hong2020language}
Yicong Hong, Cristian Rodriguez, Yuankai Qi, Qi~Wu, and Stephen Gould.
\newblock Language and visual entity relationship graph for agent navigation.
\newblock {\em Advances in Neural Information Processing Systems},
  33:7685--7696, 2020.

\bibitem{lin2021scene}
Xiangru Lin, Guanbin Li, and Yizhou Yu.
\newblock Scene-intuitive agent for remote embodied visual grounding.
\newblock In {\em Proceedings of the IEEE/CVF Conference on Computer Vision and
  Pattern Recognition}, pages 7036--7045, 2021.

\bibitem{fang2019scene}
Kuan Fang, Alexander Toshev, Li~Fei-Fei, and Silvio Savarese.
\newblock Scene memory transformer for embodied agents in long-horizon tasks.
\newblock In {\em Proceedings of the IEEE/CVF Conference on Computer Vision and
  Pattern Recognition}, pages 538--547, 2019.

\bibitem{deng2020evolving}
Zhiwei Deng, Karthik Narasimhan, and Olga Russakovsky.
\newblock Evolving graphical planner: Contextual global planning for
  vision-and-language navigation.
\newblock In {\em Advances in Neural Information Processing Systems},
  volume~33, 2020.

\bibitem{wang2021structured}
Hanqing Wang, Wenguan Wang, Wei Liang, Caiming Xiong, and Jianbing Shen.
\newblock Structured scene memory for vision-language navigation.
\newblock In {\em Proceedings of the IEEE/CVF Conference on Computer Vision and
  Pattern Recognition}, pages 8455--8464, 2021.

\bibitem{li2019robust}
Xiujun Li, Chunyuan Li, Qiaolin Xia, Yonatan Bisk, Asli Celikyilmaz, Jianfeng
  Gao, Noah~A Smith, and Yejin Choi.
\newblock Robust navigation with language pretraining and stochastic sampling.
\newblock In {\em Proceedings of the 2019 Conference on Empirical Methods in
  Natural Language Processing and the 9th International Joint Conference on
  Natural Language Processing (EMNLP-IJCNLP)}, pages 1494--1499, 2019.

\bibitem{devlin2019bert}
Jacob Devlin, Ming-Wei Chang, Kenton Lee, and Kristina Toutanova.
\newblock Bert: Pre-training of deep bidirectional transformers for language
  understanding.
\newblock In {\em Proceedings of the 2019 Conference of the North American
  Chapter of the Association for Computational Linguistics: Human Language
  Technologies, Volume 1 (Long and Short Papers)}, pages 4171--4186, 2019.

\bibitem{hao2020towards}
Weituo Hao, Chunyuan Li, Xiujun Li, Lawrence Carin, and Jianfeng Gao.
\newblock Towards learning a generic agent for vision-and-language navigation
  via pre-training.
\newblock In {\em Proceedings of the IEEE/CVF Conference on Computer Vision and
  Pattern Recognition}, pages 13137--13146, 2020.

\bibitem{ranzato2015sequence}
Marc'Aurelio Ranzato, Sumit Chopra, Michael Auli, and Wojciech Zaremba.
\newblock Sequence level training with recurrent neural networks.
\newblock {\em International Conference on Learning Representations}, 2016.

\bibitem{ross2011reduction}
St{\'e}phane Ross, Geoffrey Gordon, and Drew Bagnell.
\newblock A reduction of imitation learning and structured prediction to
  no-regret online learning.
\newblock In {\em Proceedings of the fourteenth international conference on
  artificial intelligence and statistics}, pages 627--635. JMLR Workshop and
  Conference Proceedings, 2011.

\bibitem{bengio2015scheduled}
Samy Bengio, Oriol Vinyals, Navdeep Jaitly, and Noam Shazeer.
\newblock Scheduled sampling for sequence prediction with recurrent neural
  networks.
\newblock In {\em Advances in Neural Information Processing Systems},
  volume~28, 2015.

\bibitem{sutton2018reinforcement}
Richard~S Sutton and Andrew~G Barto.
\newblock {\em Reinforcement learning: An introduction}.
\newblock MIT press, 2018.

\bibitem{parisotto2020stabilizing}
Emilio Parisotto, Francis Song, Jack Rae, Razvan Pascanu, Caglar Gulcehre,
  Siddhant Jayakumar, Max Jaderberg, Raphael~Lopez Kaufman, Aidan Clark, Seb
  Noury, et~al.
\newblock Stabilizing transformers for reinforcement learning.
\newblock In {\em International Conference on Machine Learning}, pages
  7487--7498. PMLR, 2020.

\bibitem{mnih2016asynchronous}
Volodymyr Mnih, Adria~Puigdomenech Badia, Mehdi Mirza, Alex Graves, Timothy
  Lillicrap, Tim Harley, David Silver, and Koray Kavukcuoglu.
\newblock Asynchronous methods for deep reinforcement learning.
\newblock In {\em International conference on machine learning}, pages
  1928--1937. PMLR, 2016.

\bibitem{shridhar2020alfred}
Mohit Shridhar, Jesse Thomason, Daniel Gordon, Yonatan Bisk, Winson Han,
  Roozbeh Mottaghi, Luke Zettlemoyer, and Dieter Fox.
\newblock Alfred: A benchmark for interpreting grounded instructions for
  everyday tasks.
\newblock In {\em Proceedings of the IEEE/CVF conference on computer vision and
  pattern recognition}, pages 10740--10749, 2020.

\bibitem{wang2020soft}
Hu~Wang, Qi~Wu, and Chunhua Shen.
\newblock Soft expert reward learning for vision-and-language navigation.
\newblock In {\em Computer Vision--ECCV 2020: 16th European Conference,
  Glasgow, UK, August 23--28, 2020, Proceedings, Part IX 16}, pages 126--141.
  Springer, 2020.

\bibitem{vaswani2017attention}
Ashish Vaswani, Noam Shazeer, Niki Parmar, Jakob Uszkoreit, Llion Jones,
  Aidan~N Gomez, {\L}ukasz Kaiser, and Illia Polosukhin.
\newblock Attention is all you need.
\newblock In {\em Advances in neural information processing systems}, pages
  5998--6008, 2017.

\bibitem{landi2019perceive}
Federico Landi, Lorenzo Baraldi, Marcella Cornia, Massimiliano Corsini, and
  Rita Cucchiara.
\newblock Perceive, transform, and act: Multi-modal attention networks for
  vision-and-language navigation.
\newblock {\em arXiv preprint arXiv:1911.12377}, 2019.

\bibitem{he2016deep}
Kaiming He, Xiangyu Zhang, Shaoqing Ren, and Jian Sun.
\newblock Deep residual learning for image recognition.
\newblock In {\em Proceedings of the IEEE conference on computer vision and
  pattern recognition}, pages 770--778, 2016.

\bibitem{hochreiter1997long}
Sepp Hochreiter and J{\"u}rgen Schmidhuber.
\newblock Long short-term memory.
\newblock {\em Neural computation}, 9(8):1735--1780, 1997.

\bibitem{gupta2017cognitive}
Saurabh Gupta, James Davidson, Sergey Levine, Rahul Sukthankar, and Jitendra
  Malik.
\newblock Cognitive mapping and planning for visual navigation.
\newblock In {\em Proceedings of the IEEE Conference on Computer Vision and
  Pattern Recognition}, pages 2616--2625, 2017.

\bibitem{savinov2018semi}
Nikolay Savinov, Alexey Dosovitskiy, and Vladlen Koltun.
\newblock Semi-parametric topological memory for navigation.
\newblock {\em International Conference on Learning Representations}, 2018.

\bibitem{pashevich2021episodic}
Alexander Pashevich, Cordelia Schmid, and Chen Sun.
\newblock Episodic transformer for vision-and-language navigation.
\newblock In {\em Proceedings of the IEEE/CVF International Conference on
  Computer Vision}, pages 15942--15952, 2021.

\bibitem{chen2020uniter}
Yen-Chun Chen, Linjie Li, Licheng Yu, Ahmed El~Kholy, Faisal Ahmed, Zhe Gan,
  Yu~Cheng, and Jingjing Liu.
\newblock Uniter: Universal image-text representation learning.
\newblock In {\em European Conference on Computer Vision}, pages 104--120.
  Springer, 2020.

\bibitem{li2020oscar}
Xiujun Li, Xi~Yin, Chunyuan Li, Pengchuan Zhang, Xiaowei Hu, Lei Zhang, Lijuan
  Wang, Houdong Hu, Li~Dong, Furu Wei, et~al.
\newblock Oscar: Object-semantics aligned pre-training for vision-language
  tasks.
\newblock In {\em European Conference on Computer Vision}, pages 121--137.
  Springer, 2020.

\bibitem{lu2019vilbert}
Jiasen Lu, Dhruv Batra, Devi Parikh, and Stefan Lee.
\newblock {ViLBERT}: Pretraining task-agnostic visiolinguistic representations
  for vision-and-language tasks.
\newblock In {\em Advances in Neural Information Processing Systems},
  volume~32, 2019.

\bibitem{tan2019lxmert}
Hao Tan and Mohit Bansal.
\newblock {LXMERT}: Learning cross-modality encoder representations from
  transformers.
\newblock In {\em Proceedings of the 2019 Conference on Empirical Methods in
  Natural Language Processing and the 9th International Joint Conference on
  Natural Language Processing (EMNLP-IJCNLP)}, pages 5103--5114, 2019.

\bibitem{kim2021vilt}
Wonjae Kim, Bokyung Son, and Ildoo Kim.
\newblock Vilt: Vision-and-language transformer without convolution or region
  supervision.
\newblock In {\em Proceedings of the 38th International Conference on Machine
  Learning}, pages 5583--5594, 2021.

\bibitem{dosovitskiy2020image}
Alexey Dosovitskiy, Lucas Beyer, Alexander Kolesnikov, Dirk Weissenborn,
  Xiaohua Zhai, Thomas Unterthiner, Mostafa Dehghani, Matthias Minderer, Georg
  Heigold, Sylvain Gelly, et~al.
\newblock An image is worth 16$\times$ 16 words: Transformers for image
  recognition at scale.
\newblock {\em International Conference on Learning Representations}, 2020.

\bibitem{majumdar2020improving}
Arjun Majumdar, Ayush Shrivastava, Stefan Lee, Peter Anderson, Devi Parikh, and
  Dhruv Batra.
\newblock Improving vision-and-language navigation with image-text pairs from
  the web.
\newblock In {\em European Conference on Computer Vision}, pages 259--274.
  Springer, 2020.

\bibitem{ba2016layer}
Jimmy~Lei Ba, Jamie~Ryan Kiros, and Geoffrey~E Hinton.
\newblock Layer normalization.
\newblock {\em arXiv preprint arXiv:1607.06450}, 2016.

\bibitem{arnab2021vivit}
Anurag Arnab, Mostafa Dehghani, Georg Heigold, Chen Sun, Mario Lu\v{c}i\'c, and
  Cordelia Schmid.
\newblock Vivit: A video vision transformer.
\newblock In {\em Proceedings of the IEEE/CVF International Conference on
  Computer Vision}, pages 6836--6846, 2021.

\bibitem{cao2020behind}
Jize Cao, Zhe Gan, Yu~Cheng, Licheng Yu, Yen-Chun Chen, and Jingjing Liu.
\newblock Behind the scene: Revealing the secrets of pre-trained
  vision-and-language models.
\newblock In {\em European Conference on Computer Vision}, pages 565--580.
  Springer, 2020.

\bibitem{deng2009imagenet}
Jia Deng, Wei Dong, Richard Socher, Li-Jia Li, Kai Li, and Li~Fei-Fei.
\newblock {ImageNet}: A large-scale hierarchical image database.
\newblock In {\em 2009 IEEE conference on computer vision and pattern
  recognition}, pages 248--255. {IEEE}, 2009.

\bibitem{chang2017matterport3d}
Angel Chang, Angela Dai, Thomas Funkhouser, Maciej Halber, Matthias Niebner,
  Manolis Savva, Shuran Song, Andy Zeng, and Yinda Zhang.
\newblock Matterport3d: Learning from rgb-d data in indoor environments.
\newblock In {\em 2017 International Conference on 3D Vision (3DV)}, pages
  667--676. IEEE, 2017.

\bibitem{ilharco2019general}
Gabriel Ilharco, Vihan Jain, Alexander Ku, Eugene Ie, and Jason Baldridge.
\newblock General evaluation for instruction conditioned navigation using
  dynamic time warping.
\newblock In {\em Visually Grounded Interaction and Language (ViGIL), NeurIPS
  2019 Workshop}, 2019.

\bibitem{zhu2020vision}
Fengda Zhu, Yi~Zhu, Xiaojun Chang, and Xiaodan Liang.
\newblock Vision-language navigation with self-supervised auxiliary reasoning
  tasks.
\newblock In {\em Proceedings of the IEEE/CVF Conference on Computer Vision and
  Pattern Recognition}, pages 10012--10022, 2020.

\bibitem{shrivastava2021visitron}
Ayush Shrivastava, Karthik Gopalakrishnan, Yang Liu, Robinson Piramuthu, Gokhan
  T{\"u}r, Devi Parikh, and Dilek Hakkani-T{\"u}r.
\newblock Visitron: Visual semantics-aligned interactively trained
  object-navigator.
\newblock {\em arXiv preprint arXiv:2105.11589}, 2021.

\bibitem{wang2020environment}
Xin~Eric Wang, Vihan Jain, Eugene Ie, William~Yang Wang, Zornitsa Kozareva, and
  Sujith Ravi.
\newblock Environment-agnostic multitask learning for natural language grounded
  navigation.
\newblock In {\em Computer Vision--ECCV 2020: 16th European Conference,
  Glasgow, UK, August 23--28, 2020, Proceedings, Part XXIV 16}, pages 413--430.
  Springer, 2020.

\bibitem{krantz2020beyond}
Jacob Krantz, Erik Wijmans, Arjun Majumdar, Dhruv Batra, and Stefan Lee.
\newblock Beyond the nav-graph: Vision-and-language navigation in continuous
  environments.
\newblock In {\em European Conference on Computer Vision}, pages 104--120.
  Springer, 2020.

\bibitem{gutmann2010noise}
Michael Gutmann and Aapo Hyv{\"a}rinen.
\newblock Noise-contrastive estimation: A new estimation principle for
  unnormalized statistical models.
\newblock In {\em Proceedings of the Thirteenth International Conference on
  Artificial Intelligence and Statistics}, pages 297--304. JMLR Workshop and
  Conference Proceedings, 2010.

\bibitem{loshchilov2017decoupled}
Ilya Loshchilov and Frank Hutter.
\newblock Decoupled weight decay regularization.
\newblock In {\em 7th International Conference on Learning Representations},
  2019.

\bibitem{cubuk2020randaugment}
Ekin~D Cubuk, Barret Zoph, Jonathon Shlens, and Quoc~V Le.
\newblock Randaugment: Practical automated data augmentation with a reduced
  search space.
\newblock In {\em Proceedings of the IEEE/CVF Conference on Computer Vision and
  Pattern Recognition Workshops}, pages 702--703, 2020.

\bibitem{huang2016deep}
Gao Huang, Yu~Sun, Zhuang Liu, Daniel Sedra, and Kilian~Q Weinberger.
\newblock Deep networks with stochastic depth.
\newblock In {\em European conference on computer vision}, pages 646--661.
  Springer, 2016.

\bibitem{conneau2020unsupervised}
Alexis Conneau, Kartikay Khandelwal, Naman Goyal, Vishrav Chaudhary, Guillaume
  Wenzek, Francisco Guzm{\'a}n, {\'E}douard Grave, Myle Ott, Luke Zettlemoyer,
  and Veselin Stoyanov.
\newblock Unsupervised cross-lingual representation learning at scale.
\newblock In {\em Proceedings of the 58th Annual Meeting of the Association for
  Computational Linguistics}, pages 8440--8451, 2020.

\bibitem{radford2021learning}
Alec Radford, Jong~Wook Kim, Chris Hallacy, Aditya Ramesh, Gabriel Goh,
  Sandhini Agarwal, Girish Sastry, Amanda Askell, Pamela Mishkin, Jack Clark,
  et~al.
\newblock Learning transferable visual models from natural language
  supervision.
\newblock In {\em Proceedings of the 38th International Conference on Machine
  Learning}, pages 8748--8763, 2021.

\end{thebibliography}

% \input{checklist}

%%%%%%%%%%%%%%%%%%%%%%%%%%%%%%%%%%%%%%%%%%%%%%%%%%%%%%%%%%%%
%%%%%%%%%%%%%%%%%%%%%%%%%%%%%%%%%%%%%%%%%%%%%%%%%%%%%%%%%%%%
\newpage
\appendix

\section*{Supplementary Material for HAMT}

Section~\ref{sec:model_details} provides additional details for the model. The experimental setup is described in Section~\ref{sec:expr_details}, including datasets, metrics and implementation details.
Section~\ref{sec:more_results} presents computation time on R2R dataset and full experimental results on RxR and REVERIE datasets. Section~\ref{sec:more_ablation_results} includes more ablations. 
Finally, Section~\ref{sec:viz_results} illustrates qualitative results.

\section{Model details}
\label{sec:model_details}
\subsection{Proxy tasks in training}
We employ five proxy tasks to train HAMT and introduced SAP/SAR and SPREL in Section~\ref{sec:model_pretrain}. In the following, we present the other three proxy tasks, which are all based on the input pair $(\mathcal{W}, \mathcal{H}_T)$, where $\mathcal{W}$ is the textual instruction and $\mathcal{H}_T$ is the full trajectory with length $T$.

\noindent\textbf{Masked Language Modeling (MLM).}
The task predicts masked words based on contextual words and the full trajectory.
We randomly mask out tokens in $\mathcal{W}$ with the probability of 15\% with a special token \verb|[mask]| as in BERT, and predict the word distribution $p (w_i|\mathcal{W}_{\backslash i}, \mathcal{H}_T)=f_{\text{MLM}}(x'_i)$ where $\mathcal{W}_{\backslash i}$ is the masked instruction, $x'_i$ is the output embedding of the masked word $w_i$ and $f_{\text{MLM}}$ is a two-layer fully-connected network.
The objective is to minimize the negative log-likelihood of original words: $L_{\text{MLM}} = - \mathrm{log}\ p (w_i|\mathcal{W}_{\backslash i}, \mathcal{H}_T)$.
The task is beneficial to learn grounded language representations and cross-modal alignment.

\noindent\textbf{Masked Region Modeling (MRM).}
The task aims to predict semantic labels of masked observations in the trajectory given an instruction and neighboring observations.
We zero out observations in $\mathcal{H}_T$ 15\% of the time. The target of a masked $\mathcal{O}_i$ is the class probability predicted by an image classification model pretrained on ImageNet. We use ViT-B/16~\cite{dosovitskiy2020image} in this work. 
Suppose $P_i \in \mathbb{R}^{1000}$ is the target class probability for a masked $\mathcal{O}_i$, we predict $\hat{P}_i = f_{\text{MRM}}(o'_i)$ where $o'_i$ is the output embedding of masked $\mathcal{O}_i$, and minimize the KL divergence between the two probability distributions: $L_{\text{MRM}} = - \sum_{j=1}^{1000} P_{i,j} \mathrm{log}\ \hat{P}_{i,j}$.
In order to solve the task, $o'_i$ should capture temporal continuity in the history sequence and align with relevant instructions. 

\noindent\textbf{Instruction Trajectory Matching (ITM).}
The task predicts whether a pair of instruction and trajectory is aligned.
We predict the alignment score as $s(\mathcal{W}, \mathcal{H}_T) = f_{\text{ITM}}(x'_{\mathrm{cls}} \odot h'_{\mathrm{cls}})$, where $\odot$ is element-wise multiplication and $x'_{\mathrm{cls}}, h'_{\mathrm{cls}}$ are output embeddings for the text \verb|[cls]| token and the history \verb|[cls]| token respectively.
We sample 4 negative trajectories for each positive instruction-trajectory pair during training, in which two negative trajectories are randomly selected from other positive pairs in the mini-batch, two are obtained by temporally shuffling the positive trajectory.
The objective is the Noisy Contrastive Estimation loss~\cite{gutmann2010noise}: $L_{\text{ITM}} = -  \mathrm{log}\ \frac{\exp(s(\mathcal{W}, \mathcal{H}_T))}{\exp(s(\mathcal{W}, \mathcal{H}_T)) + \sum_{k=1}^{4} \exp(s(\mathcal{W}, \mathcal{H}_{T,k}^{\text{neg}}))}$.
The model is supposed to learn cross-modal alignment and be sensitive to temporal orders of history to solve the task.

\subsection{Structure variants in fine-tuning}
\label{sec:supp_hamt_nov2t}
\begin{wrapfigure}{r}{0.5\textwidth}
	\centering
	\vspace{-4em}
	\includegraphics[width=\linewidth]{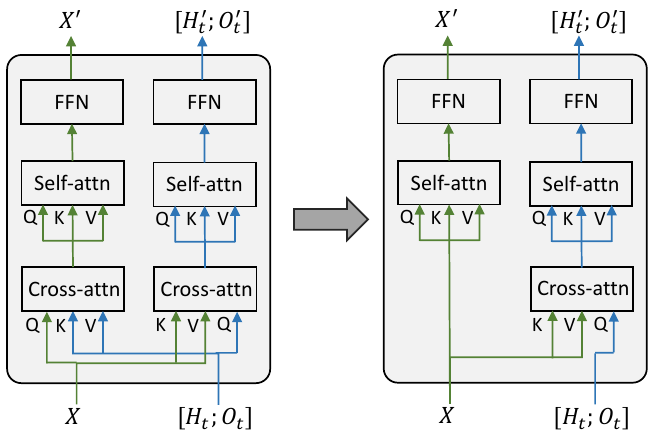}
	\caption{Comparison of the original cross-modal transformer layer (left) and the encoder-decoder based variant (right).}
	\label{fig:hamt_encdec_variant}
	\vspace{-4em}
\end{wrapfigure}

We present the encoder-decoder variant of HAMT in fine-tuning on the right of Figure~\ref{fig:hamt_encdec_variant}.
Compared to the original cross-modal transformer on the left, the variant removes text-to-vision cross-modal attention. 
The encoder encodes the texts to obtain textual embeddings. Then the decoder reuses the same text embeddings in vision-to-text attention layer at each navigation step.
In this way, the variant is more efficient when instructions are long \eg in R4R and RxR datasets.

\section{Experimental setup}
\label{sec:expr_details}
\subsection{Dataset details}
Table~\ref{tab:dataset_stats} summarizes details of the dataset split.
The proposed R2R-Back and R2R-Last setups consider exactly the same splits as the R2R dataset.
We present details to construct R2R-Back and R2R-Last in the following.

\begin{table}[h]
	\centering
	\small
	\caption{Dataset statistics. \#traj, \#instr  denote the number of trajectories and instructions respectively.}
	\label{tab:dataset_stats}
	\begin{tabular}{ccccccccc} \toprule
		\multirow{2}{*}{Dataset} & \multicolumn{2}{c}{Train} & \multicolumn{2}{c}{Val Seen} & \multicolumn{2}{c}{Val Unseen} & \multicolumn{2}{c}{Test Unseen} \\
		& \#traj & \#instr & \#traj & \#instr & \#traj & \#instr & \#traj & \#instr \\ \midrule
		R2R \cite{anderson2018vision} & 4,675 & 14,039 & 340 & 1,021 & 783 & 2,349 & 1,391 & 4,173 \\
		RxR \cite{ku2020room} & 11,077 & 79,467 & 1,244 & 8,813 & 1,517 & 13,652 & - & 11,888 \\ \midrule
		R4R \cite{jain2019stay} & 25,921 & 233,532 & 115 & 1,035 & 5,026 & 45,234 & - & - \\
		R2R-Back & 4,675 & 14,039 & 340 & 1,021 & 783 & 2,349 & - & - \\ \midrule
		CVDN \cite{thomason2020vision} & 4,742 & 4,742 & 382 & 382 & 907 & 907 & 1,384 & 1,384 \\ \midrule
		R2R-Last & 4,675 & 14,039 & 340 & 1,021 & 783 & 2,349 & - & - \\
		REVERIE \cite{qi2020reverie} & 4,150 & 10,466 & 515 & 1,423 & 1,328 & 3,521 & 2,304 & 6,292 \\ \bottomrule
	\end{tabular}
\end{table}

\noindent\textbf{R2R-Back.} We append a returning command at the end of annotated instructions in R2R to create new instructions for R2R-Back. The returning command is randomly sampled from the following sentences: ``walk back to the start'', ``return by the way you came'', ``double back to where you start'', ``backtrack to the start'', ``back the way you came'', ``return to the starting point''. The original target location is viewed as a middle stop point. The groundtruth trajectory in R2R-Back is the concatenation of the original and its inverse trajectory.

\noindent\textbf{R2R-Last.} We use spacy toolkit\footnote{\url{https://spacy.io/}} to split sentences for instructions in R2R. We only select the last sentence in each instruction as the new high-level instruction. It mainly describes where the goal location is \eg ``stop in front of the vent'', requiring the agent to explore houses without step-by-step textual guidance. The groundtruth trajectory is the same as R2R.

\subsection{Evaluation Metrics}
In R2R, RxR, R4R and R2R-Last datasets, a predicted trajectory is considered to be successful if the agent arrives 3 meters near to the final destination.
However, such definition would make a motionless agent achieve 100\% success rate (SR) on R2R-Back dataset as the final destination is the same as the starting location.
Therefore, in R2R-Back evaluation, we define the success as that an agent firstly arrives 3 meters near to the original destination and then returns 3 meters near to its starting location.
The groundtruth length in the SPL metric is also modified as the total traversed distance in groundtruth trajectory rather than the shortest distance between start and target location. 
As the REVERIE task aims for remote object grounding, the success on REVERIE is defined as arriving at a viewpoint where the target object is visible.

\subsection{Implementation Details}

\noindent\textbf{Training with proxy tasks.}
We sample proxy tasks for each mini-batch to train the HAMT model. The sampling ratio is MLM:MRM:ITM:SAP:SAR:SPREL=5:2:2:1:1:1.
The optimizer is AdamW \cite{loshchilov2017decoupled}.
In the end-to-end training stage, we use image augmentation and regularization techniques to avoid overfitting of the ViT model, including RandAugment \cite{cubuk2020randaugment} and stochastic depth \cite{huang2016deep}.

\noindent\textbf{Fine-tuning for sequential action prediction.}
Due to different goals in various VLN tasks, we design different rewards in reinforcement learning for each downstream VLN dataset.
In R2R, RxR and R4R datasets, the reward is introduced in Section~\ref{sec:model_finetune} to take both goal distance and path fidelity into account.
In R2R-Last, REVERIE and CVDN datasets where the instruction may not describe detailed navigation path, we only use the reduced distance to the goal viewpoints as rewards. We normalize the reduced distance in the same way as in the R2R dataset.
In R2R-Back dataset, we use a different fine-tune strategy to avoid trivial motionless solutions.
We require the agent to predict stop actions twice for the original destination (midpoint) and its starting point (final destination) respectively.
Before arriving at the midpoint, the RL reward is computed based on distances to the midpoint.
If the agent predicts a wrong location to stop for the midpoint, the episode is stopped; otherwise the agent continues its task while receiving rewards based on the distance to the final destination for fine-tuning.
We run each experiment twice for ablation study and use the best result on the validation unseen split for the state-of-the-art comparison.

\section{Experimental results}
\label{sec:more_results}

\subsection{Computation Efficiency}

\begin{wraptable}{r}{0.4\textwidth}
\centering
\small
\vspace{-1.5em}
\caption{Computation time in inference on R2R val unseen split.}
\label{tab:supp_r2r_time}
\begin{tabular}{lccc} \toprule
 & \begin{tabular}[c]{@{}c@{}}Inference\\ Time (s)\end{tabular} & SR & SPL \\ \midrule
RecBERT \cite{hong2020recurrent} & 69 & 63 & 57 \\
HAMT & 104 & 66 & 61 \\
HAMT noT2V & 76 & 65 & 60 \\ \bottomrule
\end{tabular}
\vspace{-1em}
\end{wraptable}

To assess the influence of history encoding on the inference time, we compare HAMT with RecBERT \cite{hong2020recurrent}. The HAMT and RecBERT use the same number of layers in the language transformer and cross-modal transformer. The main difference of two models is in the history encoding and the attended length of history for action prediction. We run each model on the R2R val unseen split (2349 instructions) and report inference times averaged over two runs using a single Tesla P100 GPU. 
For our method we compare variants with and without Text-to-Vision Attention (see Section~\ref{sec:supp_hamt_nov2t}), denoted here as HAMT and HAMT noT2V respectively.
We can see that HAMT and its noT2V variant are only 1.5x and 1.1x slower compared to RecBERT, suggesting that attending to the whole history does not increase the inference time significantly. Moreover, while HAMT noT2V is only 10\% slower compared to \cite{hong2020recurrent}, it still outperforms \cite{hong2020recurrent} in SR and SPL on val unseen split.

\subsection{RxR dataset}

\begin{wraptable}{r}{0.6\textwidth}
\centering
\small
\tabcolsep=0.1cm
\vspace{-1em}
\caption{Navigation performance on RxR test split.}
\label{tab:supp_rxr_test_cmpr}
\begin{tabular}{lccccc} \toprule
 & PL & SR$\uparrow$ & SPL$\uparrow$ & nDTW$\uparrow$ & SDTW$\uparrow$ \\ \midrule
Multilingual Baseline \cite{ku2020room} & 16.88 & 20.98 & 18.55 & 41.05 & 20.59 \\
Monolingual Baseline \cite{ku2020room} & 17.05 & 25.40 & 22.59 & 41.05 & 20.59 \\
CLIP-ViL & 15.43 & 38.34 & 35.17 & 51.10 & 32.42 \\
CLEAR-CLIP & 16.46 & 40.29 & 36.57 & 53.69 & 34.86 \\
Multilingual HAMT & 19.77 & \textbf{53.12} & \textbf{46.62} & \textbf{59.94} & \textbf{45.19} \\ \midrule
Human & 20.78 & 93.92 & 74.13 & 79.48 & 76.90 \\ \bottomrule
\end{tabular}
\end{wraptable}

As shown in Table~\ref{tab:dataset_stats}, RxR dataset contains much more instructions than R2R dataset. Therefore, we directly use RxR in training proxy tasks rather than R2R with augmented data.
As there are three different languages in RxR, we take advantage of pretrained multilingual BERT \cite{conneau2020unsupervised} to initialize the unimodal language encoder, so we are able to deal with multilingual instructions using the same HAMT model.
We employ the encoder-decoder variant of HAMT for computational efficiency.
For fair comparison with other approaches in RxR testing leaderboard\footnote{\url{https://ai.google.com/research/rxr/competition?active_tab=leaderboard} (25/10/2021).} which adopt pretrained CLIP \cite{radford2021learning} features, we use the same visual features without end-to-end optimization.
Table~\ref{tab:supp_rxr_test_cmpr} presents navigation performances on RxR test split. Our multilingual HAMT model achieves 12.83\% and 6.25\% gains on SR and nDTW respectively than the second place. Nevertheless, there is still a large gap compared to the human performance.
We further present results on val seen and val unseen splits in Table~\ref{tab:supp_rxr_val_cmpr}.

\begin{table}
\centering
\small
\caption{Navigation performances on RxR val seen and val unseen splits.}
\label{tab:supp_rxr_val_cmpr}
\begin{tabular}{lcccccccc} \toprule
 & \multicolumn{4}{c}{Val Seen} & \multicolumn{4}{c}{Val Unseen} \\
 & SR$\uparrow$ & SPL$\uparrow$ & nDTW$\uparrow$ & SDTW$\uparrow$ & SR$\uparrow$ & SPL$\uparrow$ & nDTW$\uparrow$ & SDTW$\uparrow$ \\ \midrule
Multilingual Baseline~\cite{ku2020room} & 25.2 & - & 42.2 & 20.7 & 22.8 & - & 38.9 & 18.2 \\
Monolingual Baseline~\cite{ku2020room} & 28.8 & - & 46.8 & 23.8 & 28.5 & - & 44.5 & 23.1 \\ \midrule
Multilingual HAMT & \textbf{59.4} & \textbf{58.9} & \textbf{65.3} & \textbf{50.9}  & \textbf{56.5} & \textbf{56.0} & \textbf{63.1} & \textbf{48.3} \\ \bottomrule
\end{tabular}
\end{table}

\subsection{REVERIE dataset}
The remote object localization task in REVERIE dataset requires both navigation and object grounding. To support the two subtasks in HAMT, we concatenate object features with original view image features for each viewpoint, and add an object grounding head to predict the target object given output embeddings of all object tokens.
We fine-tune HAMT that is end-to-end pretrained on R2R dataset, and use the optimized ViT to extract object features given groundtruth object bounding boxes in REVERIE dataset.
As shown in Table~\ref{tab:supp_reverie_cmpr}, HAMT achieves better navigation performance (SR and SPL), but the object grounding performance (RGS and RGSPL) on test split is worse than state of the art. 
Since HAMT can more effectively encode observed visual scenes and actions in the history sequence, it is able to better understand house environments and navigate to target viewpoints more efficiently as shown in the much higher SPL score.
However, as we use ViT optimized on R2R dataset to extract object features, the object representation might not be as generalizable as object features used in previous works which are pretrained on large-scale object detection datasets.

\begin{table}
\small
\centering
\tabcolsep=0.08cm
\caption{Navigation and object grounding performances on REVERIE val unseen and test splits.}
\label{tab:supp_reverie_cmpr}
\begin{tabular}{lcccccccccccc} \toprule
\multirow{3}{*}{Methods} & \multicolumn{6}{c}{Validation Unseen} & \multicolumn{6}{c}{Test Unseen} \\
\multicolumn{1}{c}{} & \multicolumn{4}{c}{Navigation} & \multicolumn{2}{c}{Grounding} & \multicolumn{4}{c}{Navigation} & \multicolumn{2}{c}{Grounding}  \\ 
 & TL & SR$\uparrow$ & OSR$\uparrow$ & SPL$\uparrow$ & RGS$\uparrow$ & RGSPL$\uparrow$ & TL & SR$\uparrow$ & OSR$\uparrow$ & SPL$\uparrow$ & RGS$\uparrow$ & RGSPL$\uparrow$ \\ \midrule% \midrule
%Human & - & - & - & - & - & - & 81.51 & 86.83 & 53.66 & 21.18 & 77.84 & 51.44 \\ \midrule
Seq2Seq \cite{anderson2018vision} & 11.07 & 4.20 & 8.07 & 2.84 & 2.16 & 1.63 & 10.89 & 3.99 & 6.88 & 3.09 & 2.00 & 1.58 \\
RCM \cite{wang2019reinforced} & 11.98 & 9.29 & 14.23 & 6.97 & 4.89 & 3.89  & 10.60 & 7.84 & 11.68 & 6.67 & 3.67 & 3.14 \\
SMNA \cite{ma2019self} & 9.07 & 8.15 & 11.28 & 6.44 & 4.54 & 3.61  & 9.23 & 5.80 & 8.39 & 4.53 & 3.10 & 2.39 \\
FAST-MATTN \cite{qi2020reverie} & 45.28 & 14.40 & 28.20 & 7.19 & 7.84 & 4.67 & 39.05 & 19.88 & 30.63 & 11.6 & 11.28 & 6.08 \\
SIA \cite{lin2021scene} & 41.53 & 31.53 & \textbf{44.67} & 16.28 & \textbf{22.41} & 11.56 & 48.61 & \textbf{30.80} & \textbf{44.56} & 14.85  & \textbf{19.02} & 9.20 \\
RecBERT \cite{hong2020recurrent} & 16.78 & 30.67 & 35.02 & 24.90 & 18.77 & 15.27 & 15.86 & 29.61 & 32.91 & 23.99 & 16.50 & \textbf{13.51} \\ \midrule
HAMT & 14.08 & \textbf{32.95} & 36.84 & \textbf{30.20} & 18.92 & \textbf{17.28} & 13.62 & 30.40 & 33.41 & \textbf{26.67} & 14.88  & 13.08 \\ \bottomrule
\end{tabular}
\end{table}

\section{Additional ablations}
\label{sec:more_ablation_results}
\subsection{History in training with proxy tasks}

We show that the history input plays a critical role for training with proxy tasks.
We compare HAMT with history input and PREVALENT \cite{hao2020towards} without history.
For fair comparison, we re-implement PREVALENT which only takes instruction $\mathcal{W}$ and single-step observation $\mathcal{O}_t$ as input and the other architectures are set the same as HAMT.
We train PREVALENT with all proxy tasks except the ITM task because there is no trajectory input in PREVALENT for instruction-trajectory matching.
ViT features pretrained on ImageNet are used in this experiment.

\begin{wrapfigure}{r}{0.4\textwidth}
	%\vspace{-1.2em}
	\centering
	\includegraphics[width=\linewidth]{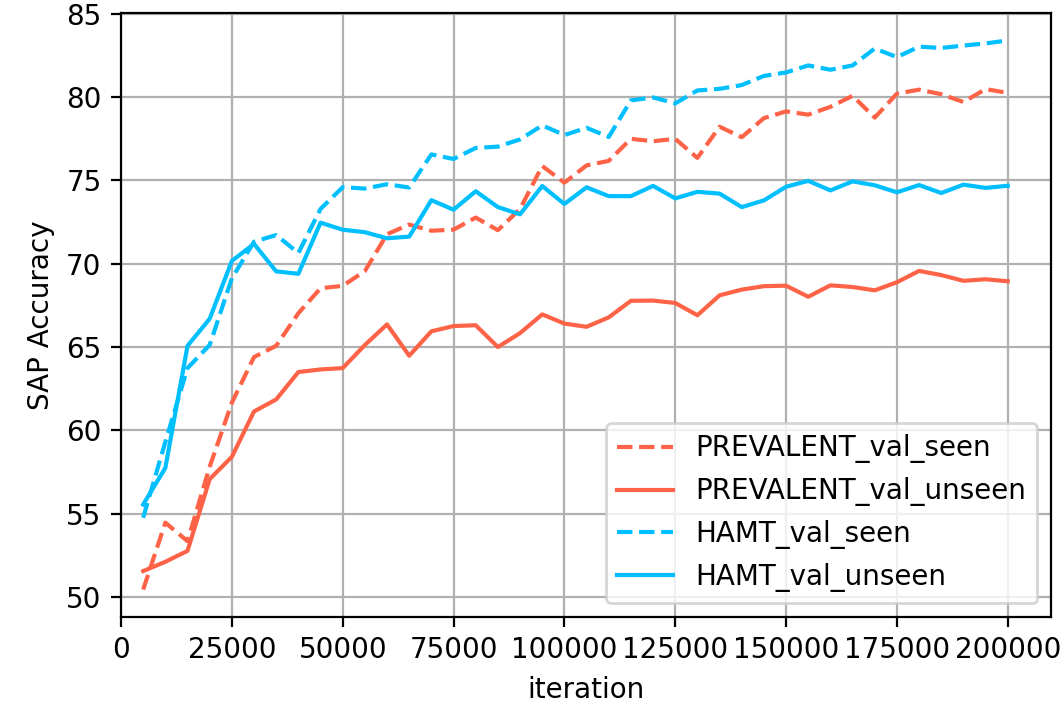}
	\caption{SAP accuracy of PREVALENT (w/o history) and HAMT (w/ history) on R2R dataset.}
	\label{fig:pretrain_sap_cmpr}
	% \vspace{-1.5em}
\end{wrapfigure}
In Figure~\ref{fig:pretrain_sap_cmpr}, we present the single-step action prediction (SAP) accuracy of HAMT and PREVALENT during the training.
The SAP accuracies on val seen split are similar for the two models, however, PREVALENT performs much worse on the val unseen split than HAMT.
Due to the capacity of large-scale transformer, PREVALENT is likely to memorize the map structure of seen houses, and thus achieves comparable performance to HAMT.
However, such knowledge cannot be transferred to unseen houses because the structure and visual observations are distinct for seen and unseen houses.
Feeding history as inputs avoids the model simply cramming the structure of seen houses, and enables it to align the history with an instruction to predict actions for better generalization.
After fine-tuning the two models on R2R dataset, we obtain SPL 57.5 on val unseen split for HAMT, while 52.7 for PREVALENT without history input.
As the same proxy tasks are used in training, the large gains of our HAMT model contribute to the history encoding.
Therefore, \textbf{the proposed history encoding can largely improve the navigation performance on top of training proxy tasks.}

\subsection{Visual features in training with proxy tasks}
\begin{wraptable}{r}{0.4\textwidth}
	\centering
	\small
	\tabcolsep=0.1cm
	\vspace{-2.5em}
    \caption{Comparison of features (same notations as Table~\ref{tab:r2r_vit_e2e}).}
    \label{tab:supp_r2r_vit_e2e}
    \begin{tabular}{ccccccc} \toprule
     &  &  & \multicolumn{2}{c}{Val Seen} & \multicolumn{2}{c}{Val Unseen} \\
    Features & PT & e2e & SR & SPL & SR & SPL \\ \midrule
    \multirow{2}{*}{\begin{tabular}[c]{@{}c@{}}Resnet \\ 152\end{tabular}} & $\times$ & $\times$ & 65.5 & 61.3 & 54.4 & 48.7 \\
    & \checkmark & $\times$ & 69.3 & 64.8 & 63.5 & 57.5 \\ \midrule
    \multirow{3}{*}{ViT} & $\times$ & $\times$ & 68.8 & 66.1 & 56.3 & 52.5 \\
     & \checkmark & $\times$ & \textbf{75.7} & \textbf{72.5} & 64.4 & 58.8 \\
     & \checkmark & \checkmark & 75.0 & 71.7 & \textbf{65.7} & \textbf{60.9} \\ \bottomrule
    \end{tabular}
	\vspace{-1em}
\end{wraptable}
Table~\ref{tab:supp_r2r_vit_e2e} provides an additional experiment in the third row compared to Table~\ref{tab:r2r_vit_e2e}.
It demonstrates that ViT features outperform ResNet152 features with and without training proxy tasks.
Comparing the last two rows in Table~\ref{tab:supp_r2r_vit_e2e}, end-to-end feature optimization improves SPL by 2.1\% on val unseen split but decreases SPL by 0.8\% on val seen split. Note that we follow previous VLN works \cite{hong2020recurrent,tan2019learning} to select the best model based on val unseen and use the same model for val seen split. We observe that the performance on val seen split can be improved with longer training time. After optimizing visual representations, HAMT converges faster on val unseen split and achieves the best performance at earlier iterations. Therefore, the performance on val seen split is slightly worse than no end-to-end optimization. If training longer, the performance with optimized ViT features on val seen split can be higher.

\subsection{Different proxy tasks in end-to-end training}
\begin{wraptable}{l}{0.4\textwidth}
\centering
\small
\vspace{-1em}
\tabcolsep=0.1cm
\caption{Comparison of different proxy tasks in end-to-end optimization.}
\label{tab:supp_proxy_e2e_ablations}
\begin{tabular}{cccccc} \toprule
 &  & \multicolumn{2}{c}{Val Seen} & \multicolumn{2}{c}{Val Unseen} \\
SAP(R) & SPREL & SR & SPL & SR & SPL \\ \midrule
$\times$ & $\times$ & 70.1 & 65.9 & 63.3 & 57.7 \\
\checkmark & $\times$ & 72.5 & 69.2 & 64.5 & 59.4 \\
\checkmark & \checkmark & \textbf{75.0} & \textbf{71.7} & \textbf{65.7} & \textbf{60.9} \\ \bottomrule
\end{tabular}
\end{wraptable}

In Table~\ref{tab:r2r_proxy_tasks} of the main paper, we fix ViT features to ablate contributions of different proxy tasks in training.
We further present the ablation results in a fully end-to-end training setup in Table~\ref{tab:supp_proxy_e2e_ablations}, where different proxy tasks are used to train HAMT including the ViT features.
The results show the same trend as Table~\ref{tab:r2r_proxy_tasks}, where our proposed two new proxy tasks (SAP/R and SPREL) are beneficial.
Moreover, we can see that the end-to-end ViT features are superior to fixed ViT features in Table~\ref{tab:r2r_proxy_tasks} on val unseen split for all the three proxy task combinations.

\subsection{Two-stage end-to-end (e2e) training strategy}
We compare our two-stage e2e training strategy with a single-stage e2e training of HAMT. 
However, single-stage e2e training achieves inferior performance to the two-stage training or even no e2e training. 
When trained for 25k iterations and evaluated on the val unseen split, the single-stage e2e training of HAMT results in SPL 53.5 while no e2e training achieves SPL 56.5. We hypothesize that the single-stage e2e training is less effective for VLN given (a) the limited training data available for the VLN task and (b) the higher complexity of VLN compared to common vision and language tasks.

\subsection{History encoding in long-horizon VLN task}

We compare different history encoding approaches on the R2R-Back dataset to show that the history information is more beneficial for the long-horizon VLN task.
Table~\ref{tab:r2rback_ablation} presents navigation results. All the models are initialized from weights after training with proxy tasks.
In order to successfully return back, the agent should remember the way it comes to the targets. The recurrent state is insufficient to capture all the information and achieves the worst navigation performance.
Encoding agent's oriented view at each step in temporal-only model improves over the recurrent approach. However, as the oriented view of the agent in backward trajectory is different from the view in forward trajectory, temporal-only model does not take advantage of the full memory in previous exploration and performs inferior to our hierarchical history encoding model.
It demonstrates the effectiveness of our proposed method in long-horizon VLN task that requires long-term dependency.
We also show that using the end-to-end trained ViT features further benefits the navigation performance.

\begin{table}[h]
	\centering
	\small
	\vspace{-1em}
	\tabcolsep=0.12cm
	\caption{Navigation results for R2R-Back dataset.}
	\label{tab:r2rback_ablation}
	\begin{tabular}{cccccccccccc} \toprule
		\multirow{2}{*}{\begin{tabular}[c]{@{}c@{}}History \\ Encoding\end{tabular}} & \multirow{2}{*}{e2e} & \multicolumn{5}{c}{Val Seen} & \multicolumn{5}{c}{Val Unseen} \\ 
		&  & TL & SR$\uparrow$ & SPL$\uparrow$ & nDTW$\uparrow$ & SDTW$\uparrow$ & TL & SR$\uparrow$ & SPL$\uparrow$ & nDTW$\uparrow$ & SDTW$\uparrow$ \\ \midrule
		Recurrent & $\times$ & 22.33 & 51.4 & 48.4 & 67.3 & 45.7 & 23.35 & 41.1 & 37.7 & 58.2 & 35.6 \\
		Temporal-only & $\times$ & 22.70 & 51.6 & 49.6 & 67.8 & 46.7 & 22.93 & 45.1 & 42.9 & 62.7 & 40.2 \\
		Hierarchical & $\times$ & 23.52 & \textbf{66.8} & \textbf{63.5} & \textbf{73.8} & \textbf{60.4} & 24.58 & 56.5 & 51.7 & 63.6 & 48.4 \\
		Hierarchical & \checkmark & \textbf{22.76} & 64.8 & 61.8 & 73.7 & 58.9 & \textbf{23.78} & \textbf{57.2} & \textbf{53.1} & \textbf{65.1} & \textbf{49.5} \\ \bottomrule
	\end{tabular}
\end{table}

\subsection{Structure variants in fine-tuning}
\begin{wraptable}{r}{0.4\linewidth}
	\centering
	\small
	\vspace{-2.5em}
	\tabcolsep=0.1cm
	\caption{Comparison of using different tokens in $f_{\text{SAP}}$ in fine-tuning.}
	\label{tab:act_token_expr}
	\begin{tabular}{ccccccc} \toprule
		\multicolumn{3}{c}{\begin{tabular}[c]{@{}c@{}}Action \\ Prediction Token\end{tabular}} & \multicolumn{2}{c}{Val Seen} & \multicolumn{2}{c}{Val Unseen} \\ \midrule
		obs & txt & hist & SR$\uparrow$ & SPL$\uparrow$ & SR$\uparrow$ & SPL$\uparrow$ \\ \midrule
		\checkmark & $\times$ & $\times$ & 76.1 & 72.8 & \textbf{66.0} & 60.3 \\
		\checkmark & \checkmark & $\times$ & 75.0 & 71.7 & 65.7 & \textbf{60.9} \\
		\checkmark & $\times$ & \checkmark & \textbf{78.0} & \textbf{75.9} & 65.5 & 60.2 \\
		\checkmark & \checkmark & \checkmark & 76.3 & 73.4 & 65.5 & 60.9 \\ \bottomrule
	\end{tabular}
	\vspace{-1em}
\end{wraptable}
Our model reuses the $f_{\text{SAP}}(o'_i \odot x'_{\text{cls}})$ in training proxy tasks to sequentially predict action in fine-tuning.
In Table~\ref{tab:act_token_expr}, we compare using different input tokens for the action prediction in $f_{\text{SAP}}$, including different combinations of the observation token $o'_i$, global history token $h'_{\text{cls}}$ and special text token $x'_{\text{cls}}$.
We can see that the performance varies little on the val unseen split, which indicates that the cross-modal transformer in our model is able to effectively fuse different modalities so that the performance is influenced little by tokens used in prediction.

% \begin{wraptable}{l}{0.4\linewidth}
% 	\centering
% 	\small
% 	\vspace{-1em}
% 	\tabcolsep=0.1cm
% 	\caption{Ablation of using text-to-vision attention in fine-tuning.}
% 	\label{tab:lang_ca_expr}
% 	\begin{tabular}{ccccc} \toprule
% 		\multirow{2}{*}{\begin{tabular}[c]{@{}c@{}}Text-to-Vision\\ Attention\end{tabular}}& \multicolumn{2}{c}{Val Seen} & \multicolumn{2}{c}{Val Unseen} \\ 
% 		& SR$\uparrow$ & SPL$\uparrow$ & SR$\uparrow$ & SPL$\uparrow$ \\ \midrule
% 		$\times$ & \textbf{77.5} & \textbf{74.0} & 64.8 & 59.8 \\
% 		\checkmark  & 75.0 & 71.7 & \textbf{65.7} & \textbf{60.9} \\ \bottomrule
% 	\end{tabular}
% % 	\vspace{-1em}
% \end{wraptable}

% We investigate the impact of text-to-vision attention in fine-tuning in Table~\ref{tab:lang_ca_expr}.
% The model without text-to-vision attention performs worse than model with the attention on unseen split by almost 1\% on SR and SPL metrics.
% But it is able to run 1.5$\times$ faster with less GPU memory footprint. Therefore, removing text-to-vision attention can balance the navigation performance and computation efficiency.

\section{Qualitative results}
\label{sec:viz_results}

\begin{figure}
	\centering
	\begin{subfigure}[b]{0.48\textwidth}
		\centering
		\includegraphics[width=\textwidth]{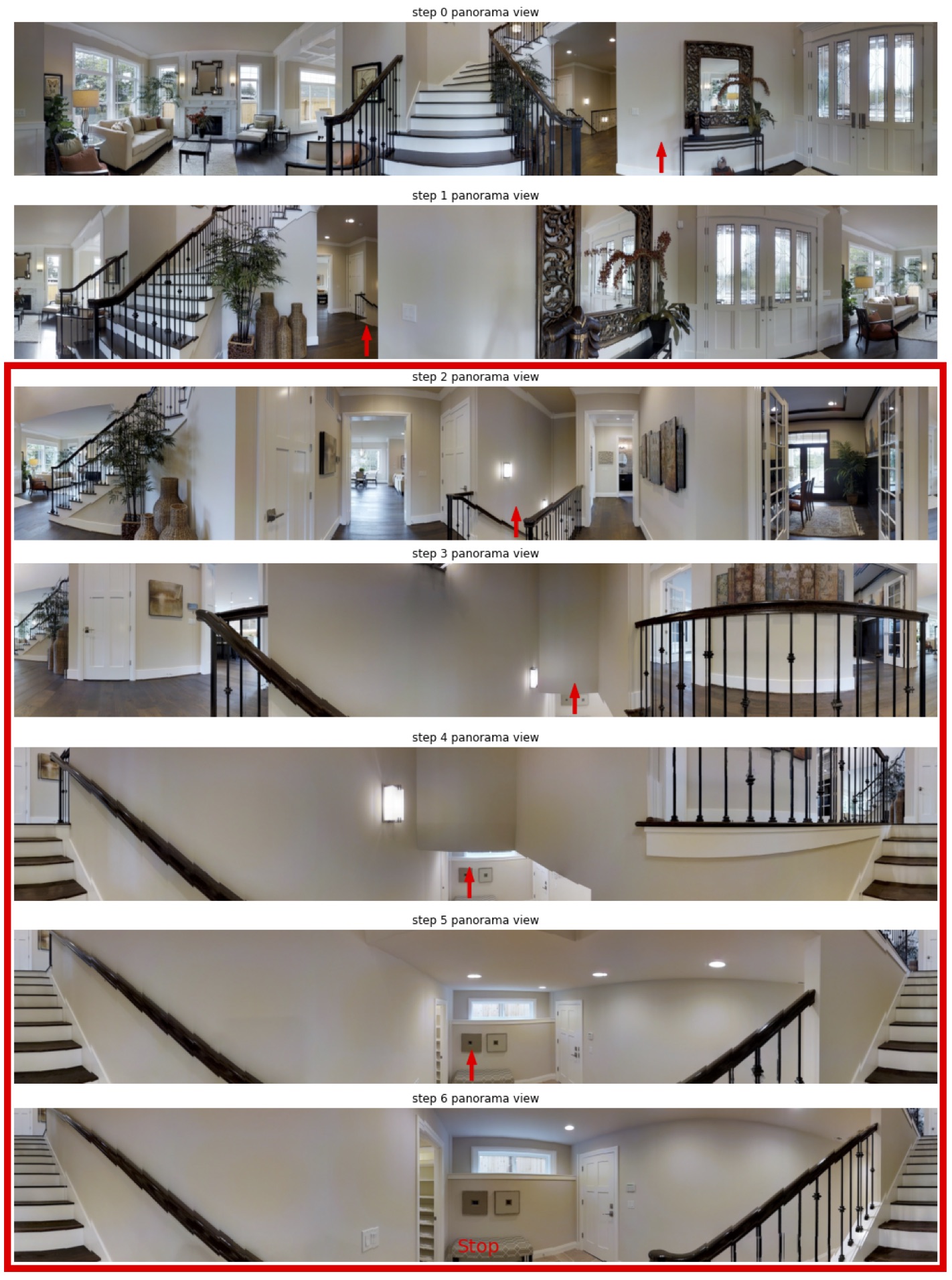}
		\caption{Predicted trajectory by RecBERT~\cite{hong2020recurrent} (failed).}
		\label{fig:ex_02_rec}
	\end{subfigure}
	\hfill
	\begin{subfigure}[b]{0.48\textwidth}
		\centering
		\includegraphics[width=\textwidth]{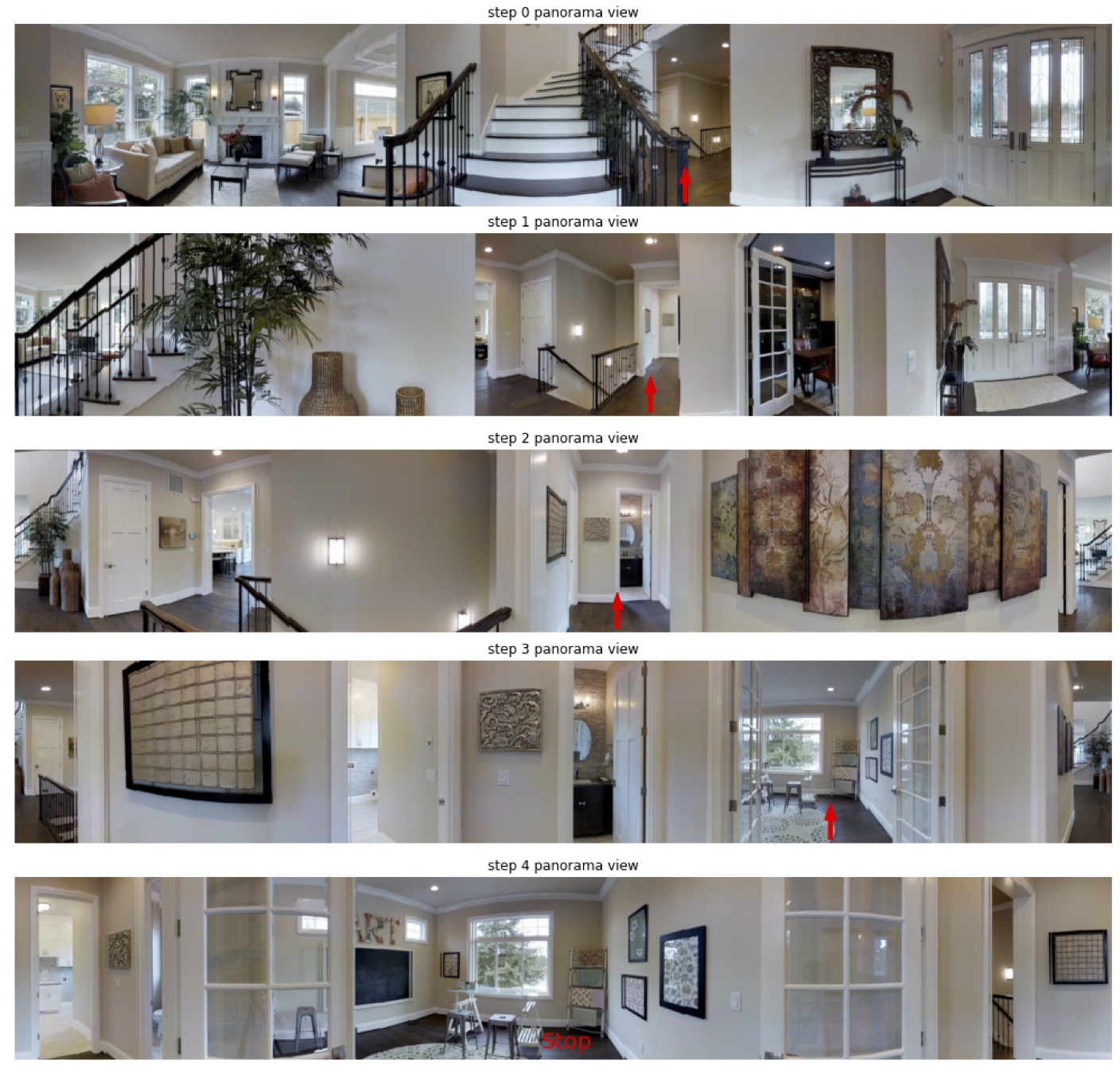}
		\caption{Predicted trajectory by HAMT (succeed).}
		\label{fig:ex_02_hamt}
	\end{subfigure}
	\caption{Examples in R2R val unseen split. Navigation steps inside red box are incorrect. The instruction is ``Walk to the right of the stairs. Continue past and to the right of the stairs that go down. Turn right and stop in the doorway of the double glass doors.'' (id: 697\_0). The RecBERT misunderstands the instruction and goes down the stairs instead of turning right. Our HAMT is better to understand the instruction and spatial relation related to the stairs to turn to the right of the stairs.}
	\label{fig:ex_02}
\end{figure}

\begin{figure}
	\centering
	\begin{subfigure}[b]{0.48\textwidth}
		\centering
		\includegraphics[width=\textwidth]{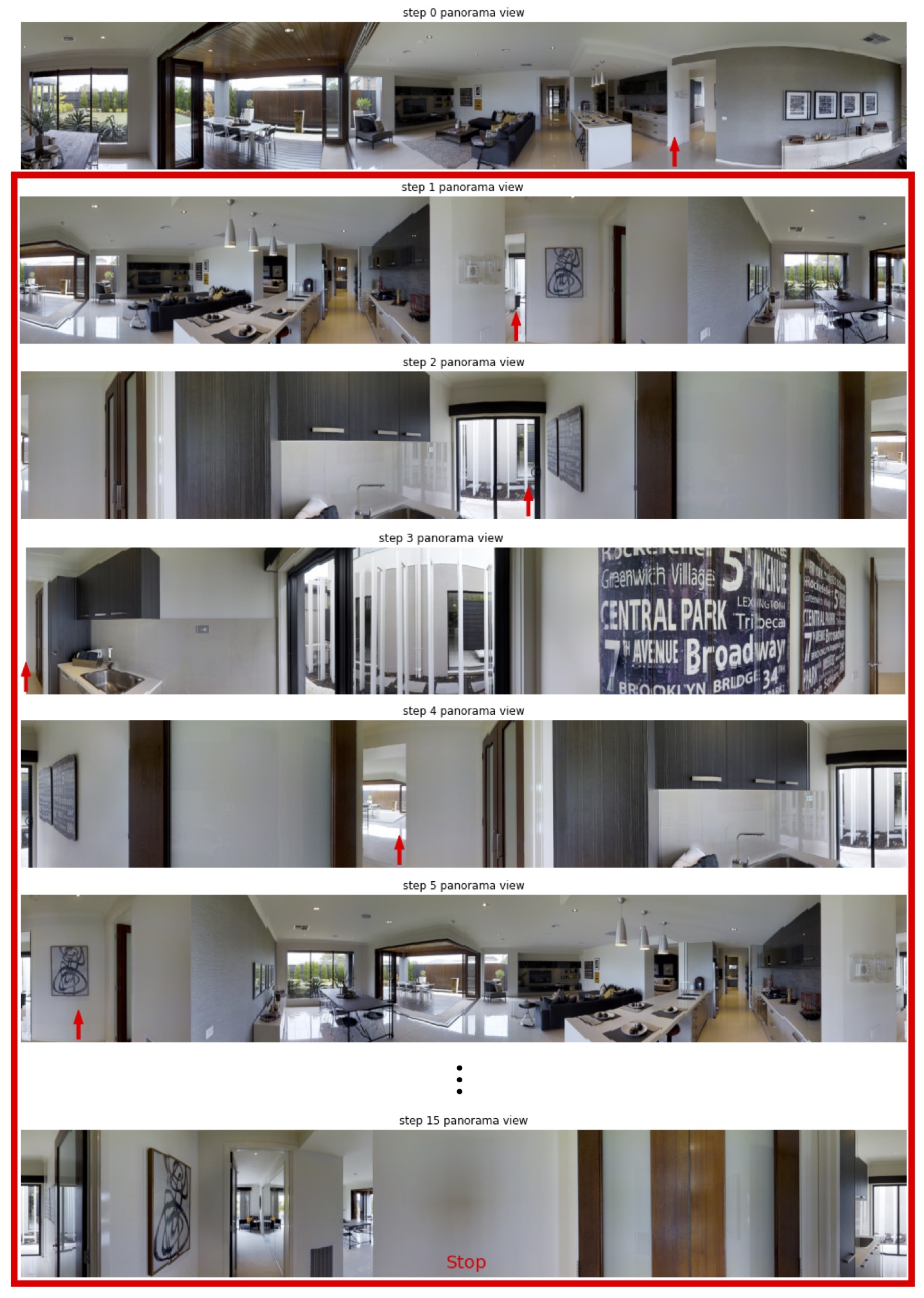}
		\caption{Predicted trajectory by RecBERT~\cite{hong2020recurrent} (failed).}
		\label{fig:ex_03_rec}
	\end{subfigure}
	\hfill
	\begin{subfigure}[b]{0.48\textwidth}
		\centering
		\includegraphics[width=\textwidth]{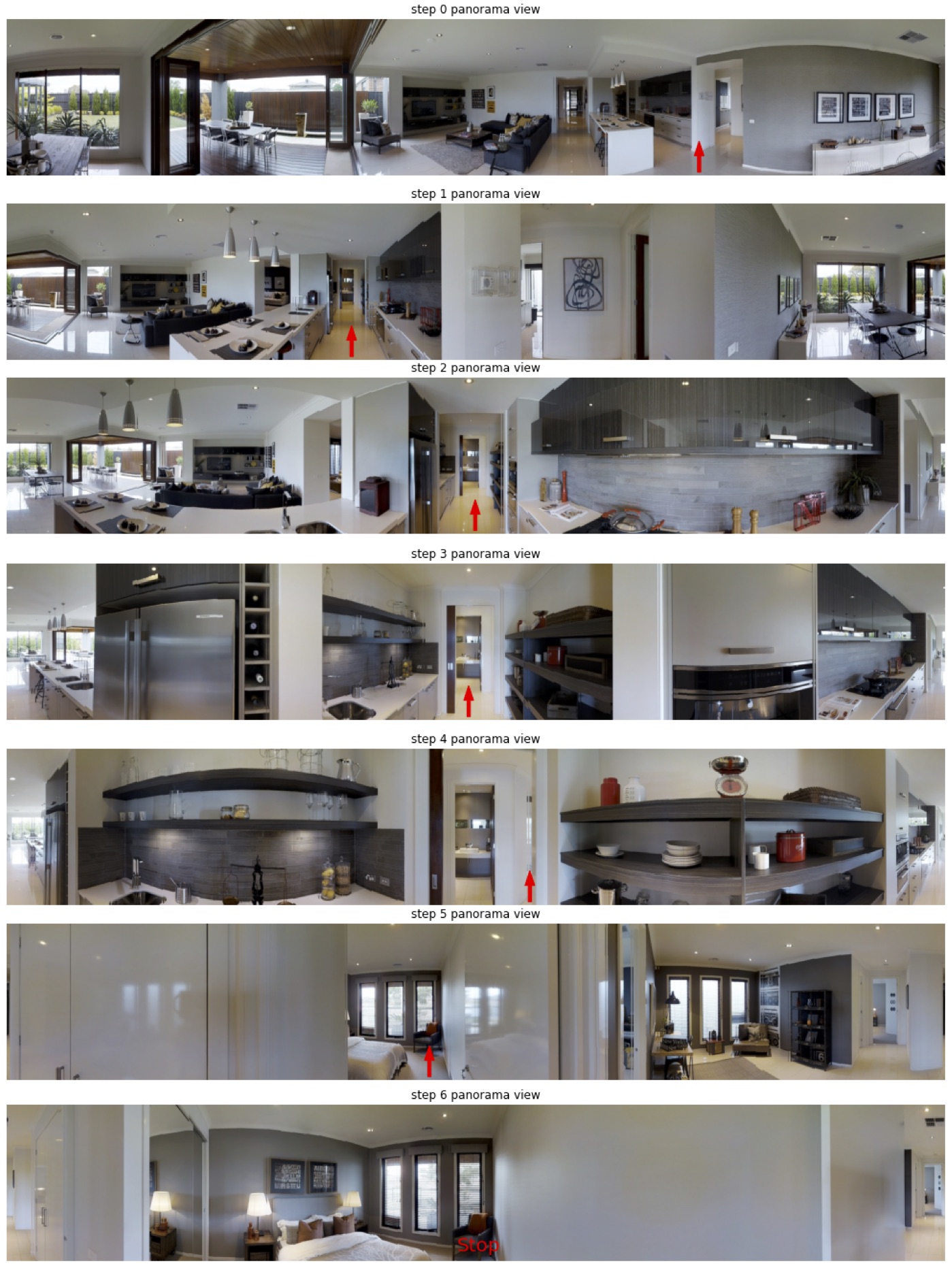}
		\caption{Predicted trajectory by HAMT (succeed).}
		\label{fig:ex_03_hamt}
	\end{subfigure}
	\caption{Examples in R2R val unseen split. Navigation steps inside red box are incorrect. The instruction is ``Walk into the kitchen area. Walk by the sink and oven. Walk straight into the hallway. Turn right into the little room. Turn left and walk into the bedroom. Stop by the corner of the bed.'' (id: 155\_0). The RecBERT fails to recognize the kitchen area and navigates back and forth in wrong locations. Our HAMT correctly recognizes the kitchen and follows the instruction.}
	\label{fig:ex_03}
\end{figure}

\begin{figure}
	\centering
	\begin{subfigure}[b]{0.48\textwidth}
		\centering
		\includegraphics[width=\textwidth]{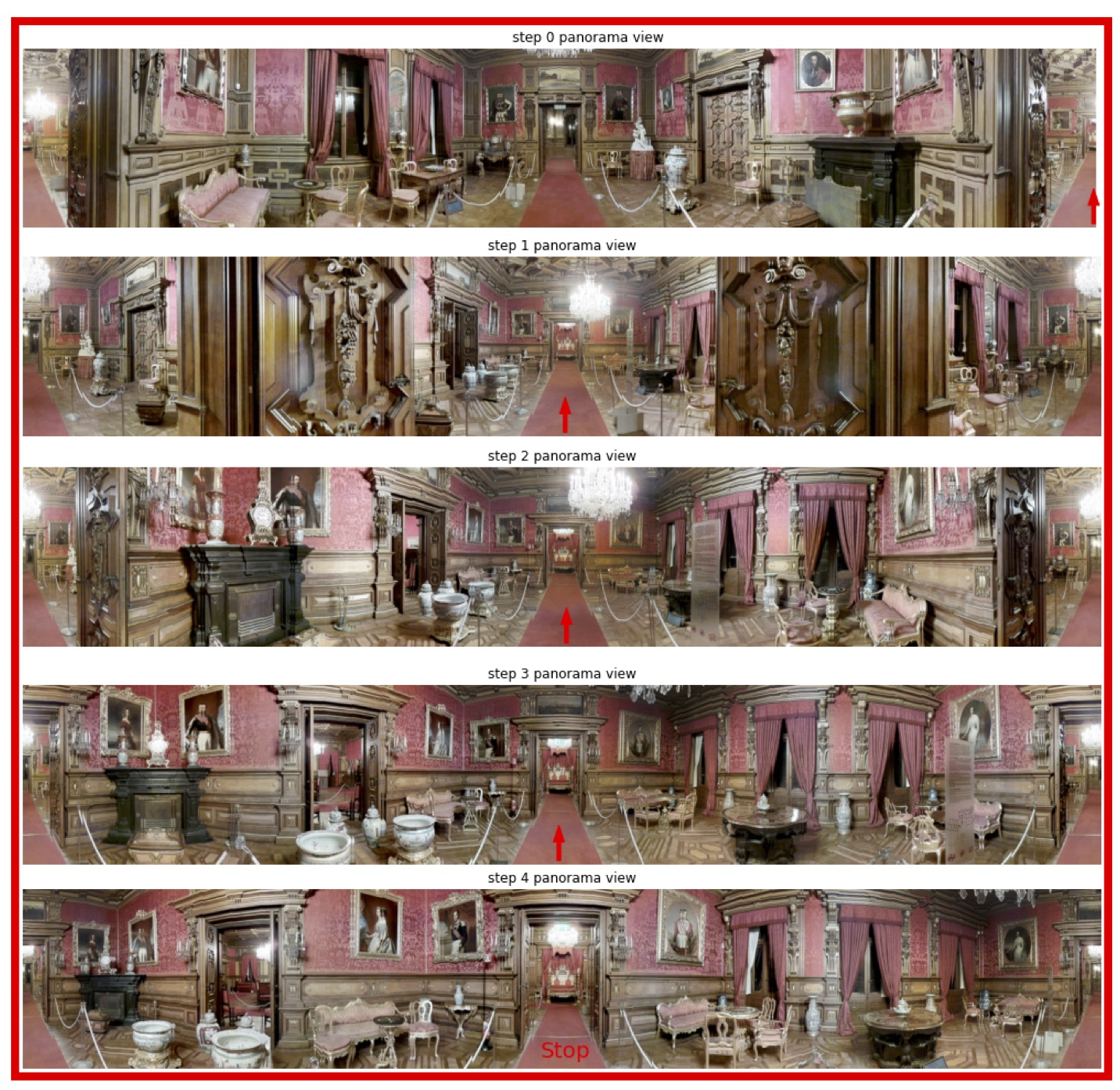}
		\caption{Predicted trajectory by RecBERT~\cite{hong2020recurrent} (failed).}
		\label{fig:ex_04_rec}
	\end{subfigure}
	\hfill
	\begin{subfigure}[b]{0.48\textwidth}
		\centering
		\includegraphics[width=\textwidth]{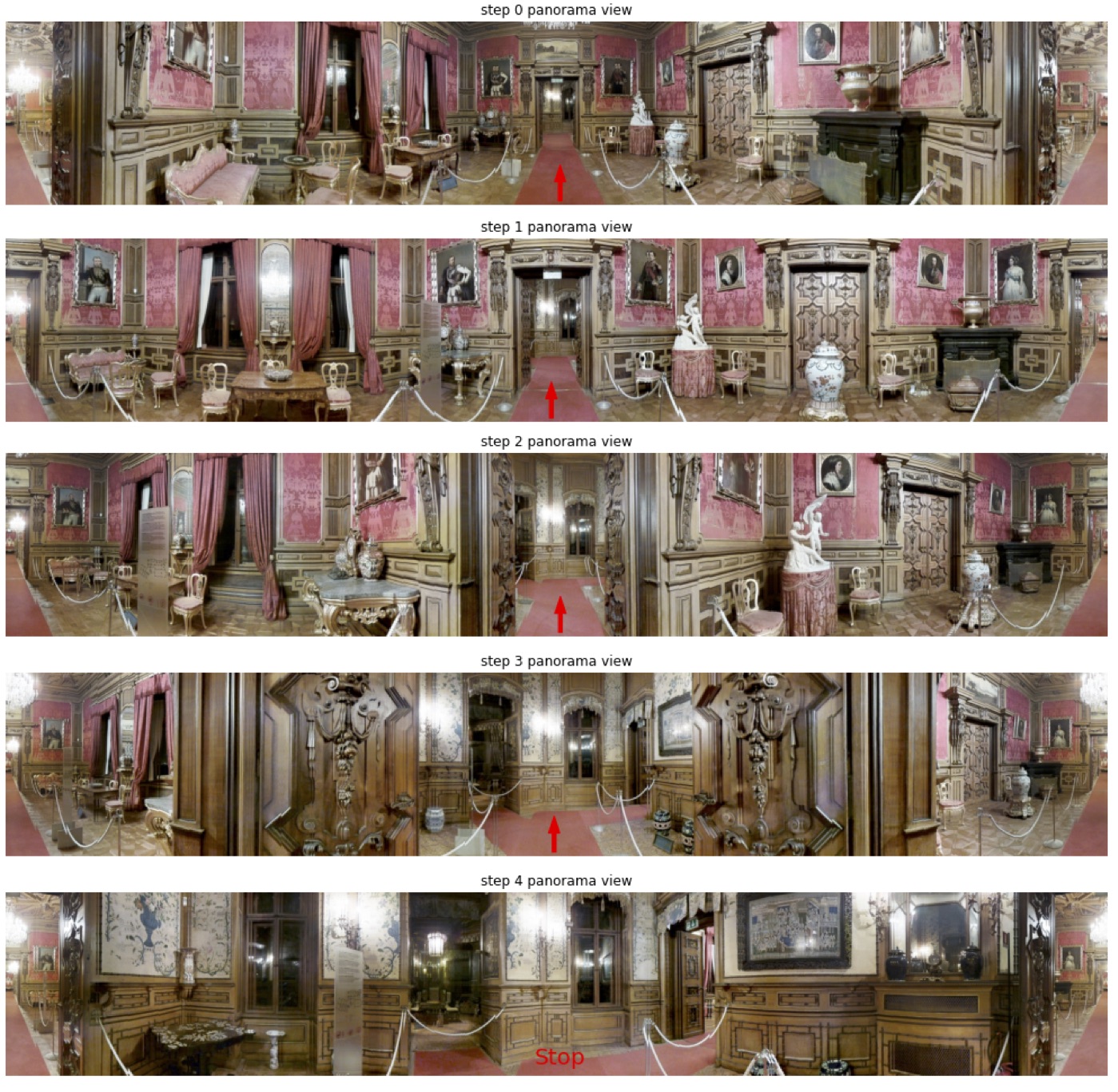}
		\caption{Predicted trajectory by HAMT (succeed).}
		\label{fig:ex_04_hamt}
	\end{subfigure}
	\caption{Examples in R2R val unseen split. Navigation steps inside red box are incorrect. The instruction is ``Walk straight until you get to a room that has a black table on the left with flowers on it. Wait there.'' (id: 4182\_2). The RecBERT takes the wrong direction at the first step, while our HAMT follows the instruction and successfully stops.}
	\label{fig:ex_04}
\end{figure}

\begin{figure}
	\centering
	\begin{subfigure}[b]{0.48\textwidth}
		\centering
		\includegraphics[width=\textwidth]{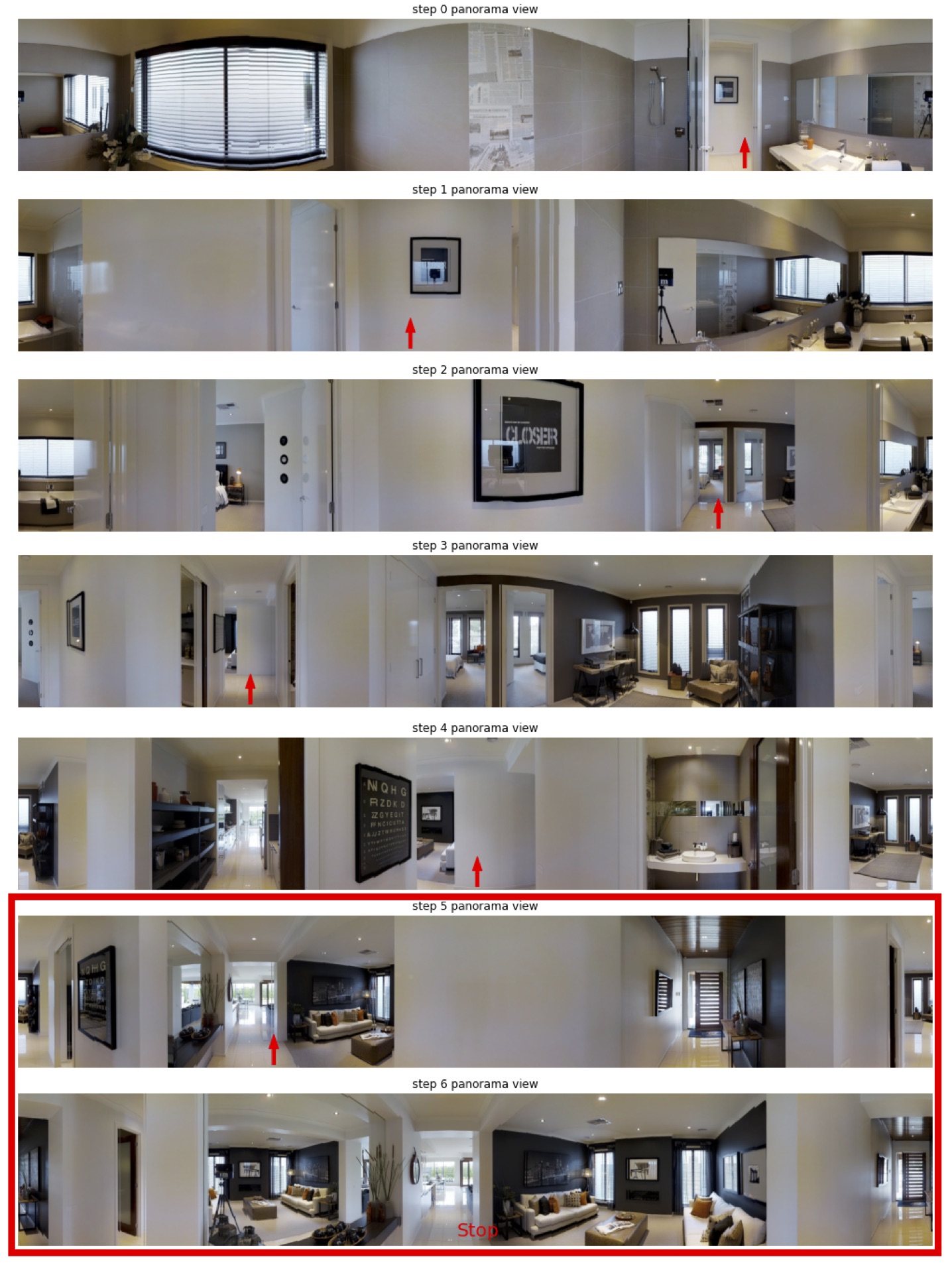}
		\caption{Predicted trajectory by RecBERT~\cite{hong2020recurrent} (failed).}
		\label{fig:ex_01_rec}
	\end{subfigure}
	\hfill
	\begin{subfigure}[b]{0.48\textwidth}
		\centering
		\includegraphics[width=\textwidth]{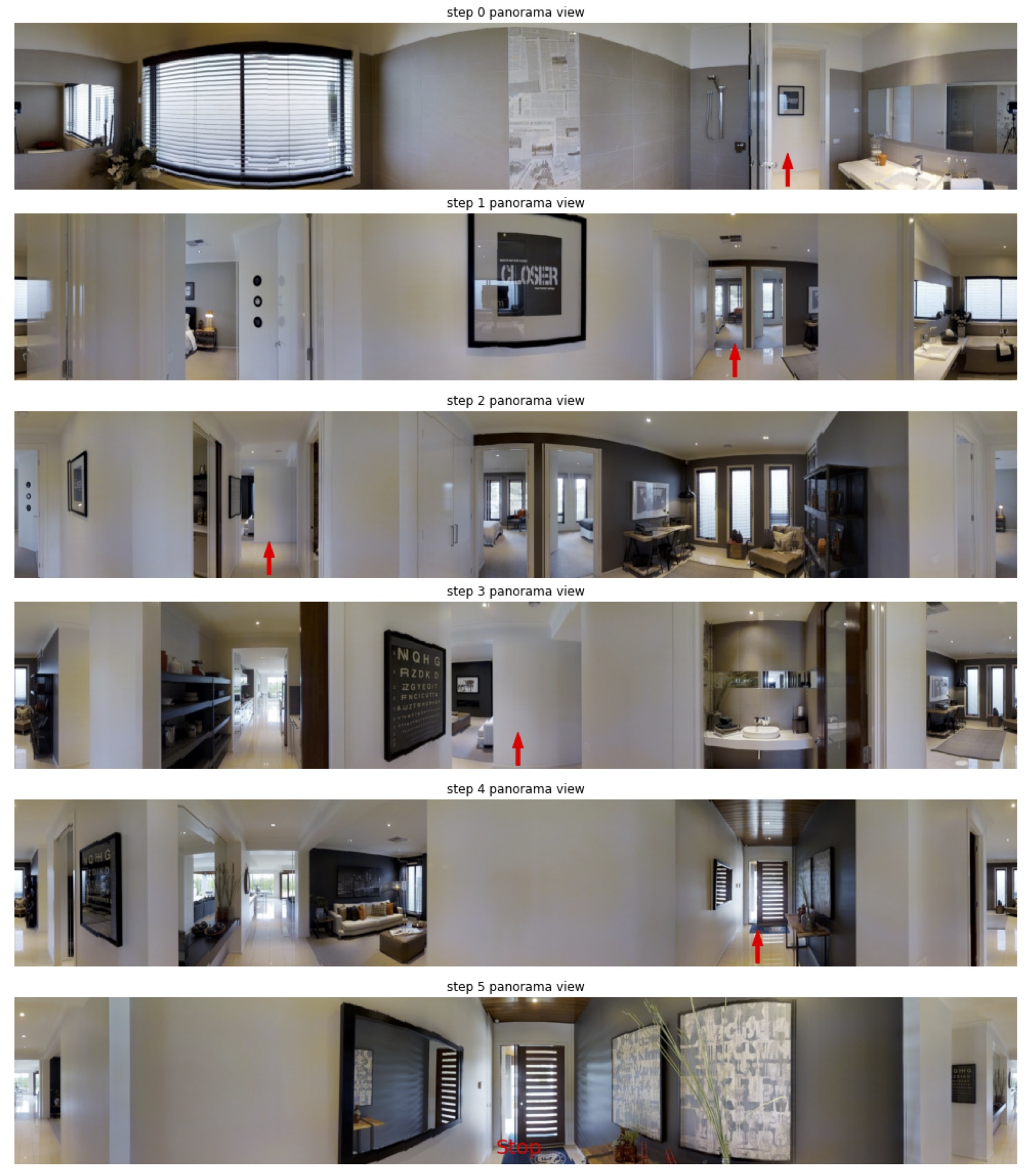}
		\caption{Predicted trajectory by HAMT (succeed).}
		\label{fig:ex_01_hamt}
	\end{subfigure}
	\caption{Examples in R2R val unseen split. Navigation steps inside red box are incorrect. The instruction is ``Walk out of the bathroom and turn right. Turn left and walk down the hallway. Turn right and stop by the end table.'' (id: 5153\_0). The RecBERT correctly performs the first two turns but fails to track the third turn right action and stops incorrectly. Our HAMT is better to align the current state with the instruction to correctly perform  the third turn right action.}
	\label{fig:ex_01}
\end{figure}

\begin{figure}
	\centering
	\begin{subfigure}[b]{0.48\textwidth}
		\centering
		\includegraphics[width=\textwidth]{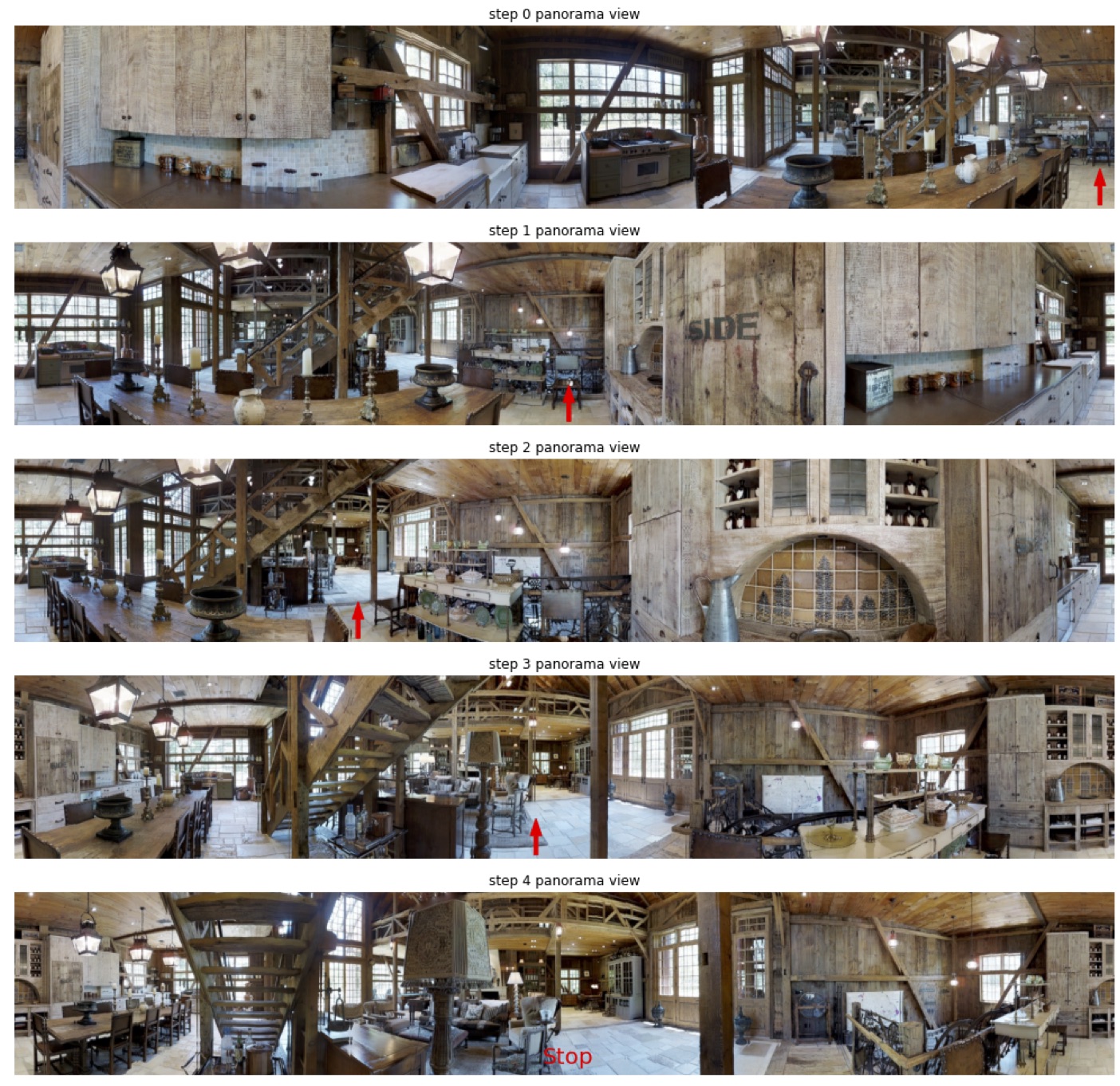}
		\caption{groundtruth trajectory.}
		\label{fig:fail_02_gt}
	\end{subfigure}
	\hfill
	\begin{subfigure}[b]{0.48\textwidth}
		\centering
		\includegraphics[width=\textwidth]{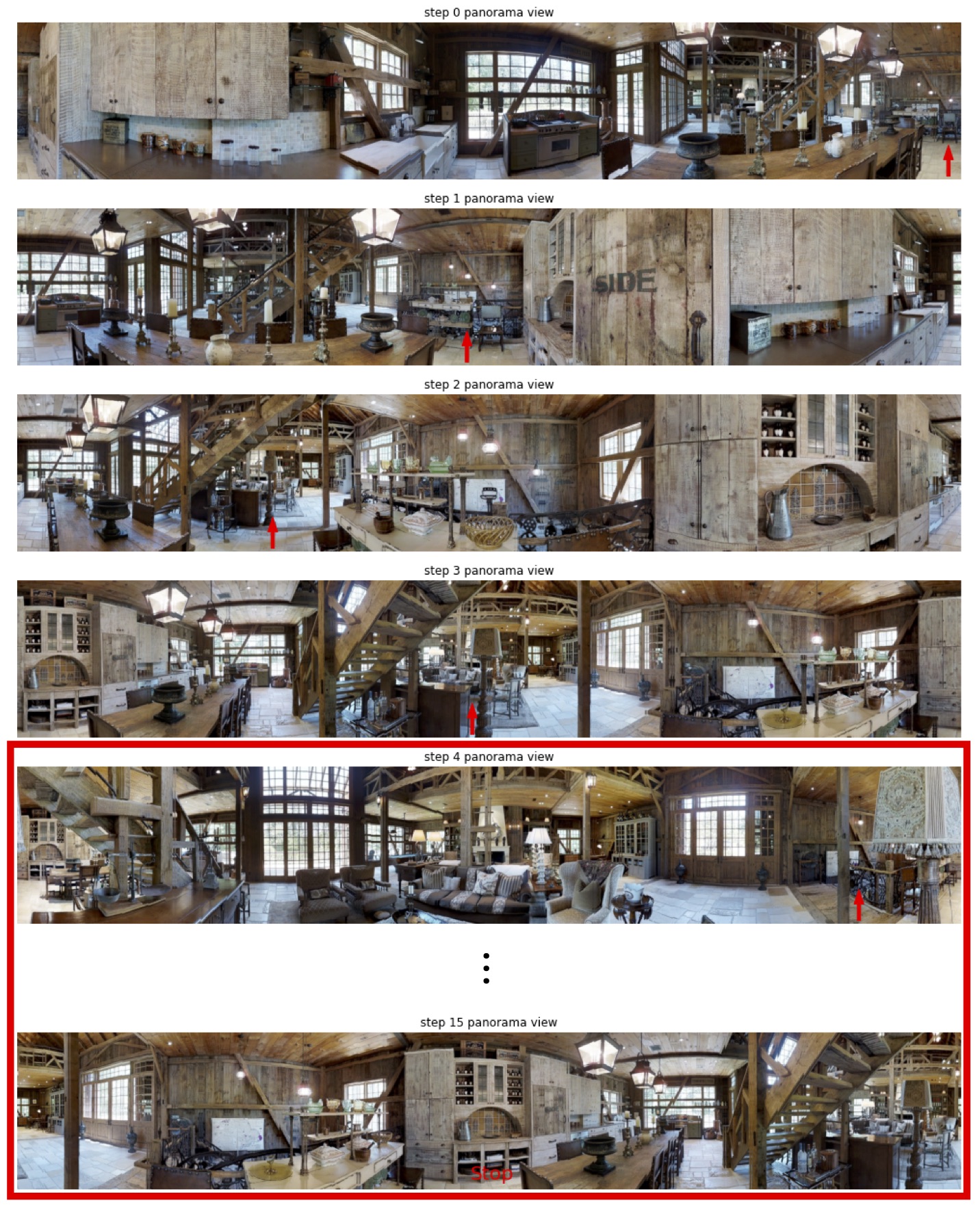}
		\caption{Predicted trajectory by HAMT (failed).}
		\label{fig:fail_02_hamt}
	\end{subfigure}
	\caption{Failure cases in R2R val unseen split. The instruction is ``Go stand underneath the stairs, next to the liquor shelf. '' (id: 36968\_2). Though HAMT correctly goes towards the direction, it fails to recognize the liquor shelf and results in exploring further the room until reaching the maximum number of navigation steps.}
	\label{fig:fail_02}
\end{figure}

\begin{figure}
	\centering
	\begin{subfigure}[b]{0.48\textwidth}
		\centering
		\includegraphics[width=\textwidth]{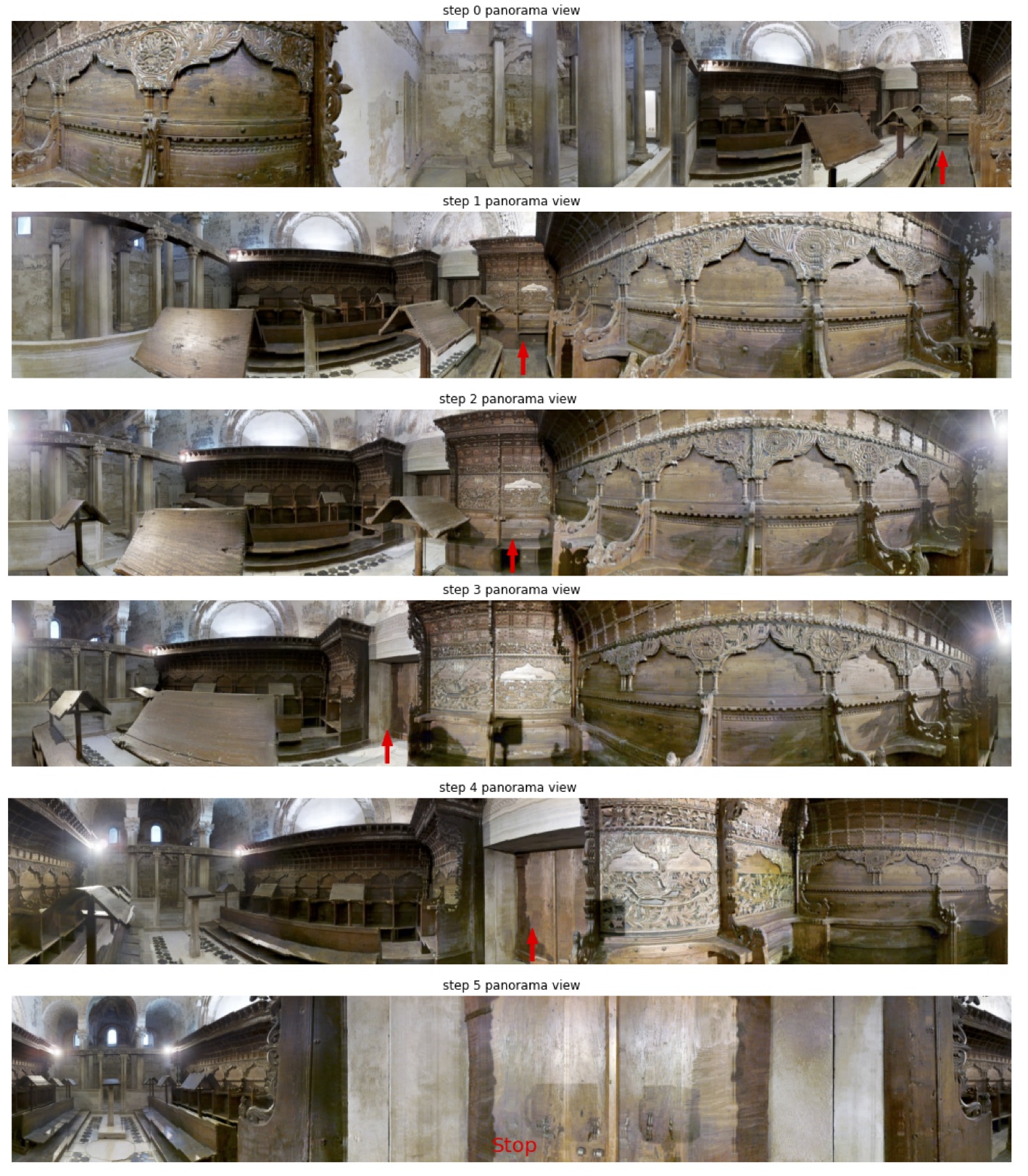}
		\caption{Groundtruth trajectory.}
		\label{fig:fail_01_gt}
	\end{subfigure}
	\hfill
	\begin{subfigure}[b]{0.48\textwidth}
		\centering
		\includegraphics[width=\textwidth]{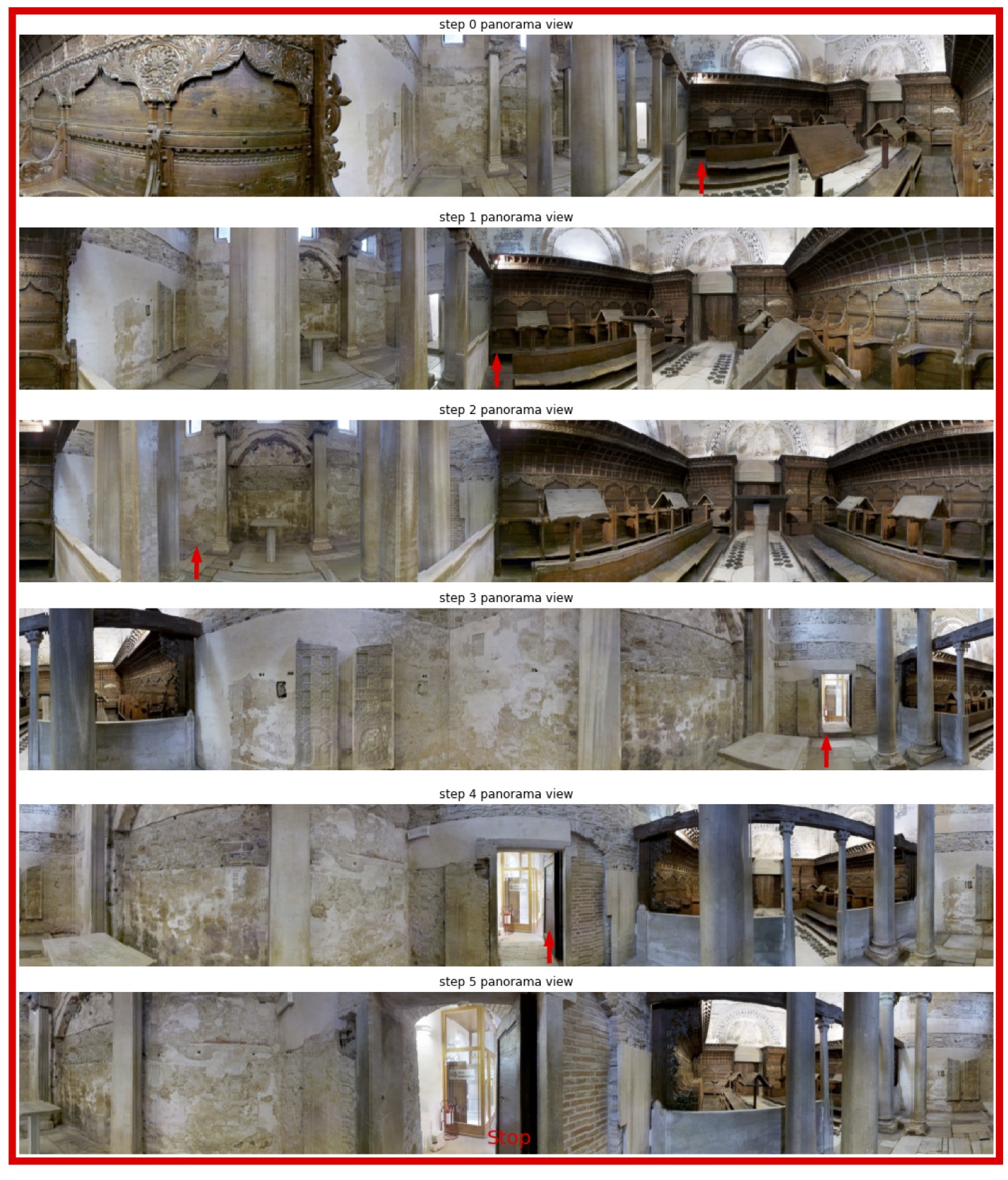}
		\caption{Predicted trajectory by HAMT (failed).}
		\label{fig:fail_01_hamt}
	\end{subfigure}
	\caption{Failure cases in R2R val unseen split. The instruction is ``With the low stone or concrete barrier behind you, walk parallel to the board covering the floor and turn left before reaching the end. Move forward to leave the wooden flooring and when on the stone flooring, turn right and stand in front of the doors leading out of the room.'' (id: 5873\_1). As the scene is unusual, HAMT fails to locate itself in the correct direction at the first step.}
	\label{fig:fail_01}
\end{figure}

Figures~\ref{fig:ex_02}-\ref{fig:ex_01} illustrate trajectories obtained by our HAMT model and compare them to results of the state-of-the-art RecBERT~\cite{hong2020recurrent} model.
We can see that HAMT enables to better interpret instructions (Figure~\ref{fig:ex_02}), recognize the scene (Figure~\ref{fig:ex_03}), follow the correct direction (Figure~\ref{fig:ex_04}), and align the current observation with the instruction (Figure~\ref{fig:ex_01}). We also provide some failure cases in Figures~\ref{fig:fail_02}-\ref{fig:fail_01}, where the HAMT model still needs improvements on scene and object recognition.

\end{document}